\documentclass{pediamed-ai}

\usepackage{amsmath,amsfonts,amssymb}
\usepackage{bm}
\usepackage{nicefrac}
\usepackage{extarrows}
\usepackage{algorithm,algpseudocode}

\usepackage{array}
\usepackage{tabularx}
\usepackage{ltablex}
\usepackage{longtable}
\usepackage{tabularray}
\usepackage{makecell}
\usepackage{diagbox}
\usepackage{hhline}
\usepackage{colortbl}   
\usepackage{adjustbox}
\usepackage{siunitx}

\usepackage{epsfig}
\usepackage{wrapfig}
\usepackage{needspace}   
\usepackage{floatflt}
\usepackage{marginnote}

\usepackage{xspace}
\usepackage{soul}
\usepackage{multicol}
\usepackage{changepage}
\usepackage{setspace}
\usepackage{enumerate}
\usepackage{enumitem}
\setlist[itemize]{leftmargin=*}
\usepackage{pifont}
\usepackage{fontawesome5}
\usepackage{wasysym}
\usepackage{titlecaps}
\usepackage{calc}
\usepackage{etoolbox}
\usepackage{scalefnt}
\usepackage{lipsum}       

\usepackage{fancyvrb}
\usepackage{fancyhdr}     
\usepackage{url}
\usepackage{titletoc}
\usepackage{tikz}

\crefname{appendix}{Appendix}{Appendices}
\crefname{section}{Section}{Sections}
\crefname{figure}{Fig.}{Figs.}
\crefname{table}{Tab.}{Tabs.}

\usepackage[symbol]{footmisc}

\renewcommand{\paragraph}[1]{%
  \noindent\textbf{#1.}\noindent\xspace%
}

\titlecontents{section}
  [1.5em]
  {\addvspace{-0.5pt}}
  {\bfseries\contentslabel{2.3em}}
  {\hspace*{-2.3em}\bfseries}
  {\bfseries\titlerule*[.5pc]{.}\contentspage}
\titlecontents{subsection}
  [3.8em]
  {\addvspace{-2.2pt}}
  {\contentslabel{2.3em}}
  {\hspace*{-2.3em}}
  {\titlerule*[.5pc]{.}\contentspage}

\usepackage{amsmath,amsfonts,bm}

\def\eqref#1{equation~\ref{#1}}

\def\1{\bm{1}}

\DeclareMathAlphabet{\mathsfit}{\encodingdefault}{\sfdefault}{m}{sl}
\SetMathAlphabet{\mathsfit}{bold}{\encodingdefault}{\sfdefault}{bx}{n}

\definecolor{logoRed}{HTML}{CB2F10}
\definecolor{logoBlue}{HTML}{1A4E8A}
\definecolor{logoCyan}{HTML}{56BBCC}

\definecolor{MyBlue}{rgb}{0.46, 0.50, 0.61}
\definecolor{MyDarkBlue}{rgb}{0, 0.08, 0.8}
\definecolor{MyDarkGreen}{RGB}{45,155,45}
\definecolor{MyDarkRed}{rgb}{0.8, 0.02, 0.02}
\definecolor{MyOrange}{rgb}{1.0, 0.4, 0.2}
\definecolor{MyPurple}{RGB}{111,0,255}
\definecolor{MyRed}{rgb}{0.8, 0.0, 0.0}
\definecolor{MyGold}{rgb}{0.75, 0.6, 0.12}
\definecolor{MyDarkgray}{rgb}{0.66, 0.66, 0.66}
\definecolor{MyBrown}{rgb}{0.65, 0.16, 0.16}
\definecolor{MyMutedRose}{rgb}{0.58, 0.29, 0.35}

\definecolor{bggray}{HTML}{F5F5F5}
\definecolor{lightgray}{rgb}{0.95, 0.95, 0.95}
\definecolor{mybluetitle}{HTML}{4B527E}
\definecolor{qcolor}{HTML}{536872}
\definecolor{applegreen}{rgb}{0.55, 0.71, 0.0}

\colorlet{titleblue}{blue!80!black}
\colorlet{titlered}{red!80!black}
\colorlet{titlegreen}{green!80!black}
\colorlet{darkgreen}{green!50!black}

\newcommand{\ignore}[1]{}

\makeatletter
\DeclareRobustCommand\onedot{\futurelet\@let@token\@onedot}
\def\@onedot{\ifx\@let@token.\else.\null\fi\xspace}

\makeatother

\newcolumntype{C}{>{\centering\arraybackslash}X}
\newcolumntype{Y}{>{\centering\arraybackslash}X}
\newcolumntype{P}[1]{>{\centering\arraybackslash}p{#1}}
\newcolumntype{L}[1]{>{\raggedright\let\newline\\\arraybackslash\hspace{0pt}}m{#1}}
\newcolumntype{R}[1]{>{\raggedleft\let\newline\\\arraybackslash\hspace{0pt}}m{#1}}
\newcolumntype{x}[1]{>{\centering\arraybackslash}p{#1pt}}
\newcolumntype{y}[1]{>{\raggedright\arraybackslash}p{#1pt}}
\newcolumntype{z}[1]{>{\raggedleft\arraybackslash}p{#1pt}}

\newlength\savewidth

\newcommand{\kindatiny}{\fontsize{6pt}{7.2pt}\selectfont}
\newcommand{\tablestyle}[2]{%
  \fontfamily{ptm}\selectfont%
  \let\itold\it%
  \def\it{\itold \fontfamily{ptm}\selectfont}%
  \setlength{\tabcolsep}{#1}\renewcommand{\arraystretch}{#2}\centering\kindatiny%
  \let\citeold\cite%
  \renewcommand{\cite}[1]{\normalfont\fontfamily{ptm}\selectfont\tiny\citeold{##1}}%
}

\newdimen\abovecrulesep \abovecrulesep=0pt
\newdimen\belowcrulesep \belowcrulesep=0pt
\makeatletter
\patchcmd{\@@@cmidrule}{\aboverulesep}{\abovecrulesep}{}{}
\patchcmd{\@xcmidrule}{\belowrulesep}{\belowcrulesep}{}{}
\makeatother

\tcbset{colback=white}

\newtcolorbox{planbox}[1]{
  colback=gray!5,
  colframe=gray!75,
  title=#1,
  fonttitle=\bfseries
}

\tcbset{
  aibox/.style={
    width=\linewidth,
    top=10pt,
    colback=white,
    colframe=black,
    colbacktitle=black,
    enhanced,
    center,
    attach boxed title to top left={yshift=-0.1in,xshift=0.15in},
    boxed title style={boxrule=0pt,colframe=white,},
  }
}
\newtcolorbox{AIbox}[2][]{aibox,title=#2,#1}

\definecolor{backcolour}{rgb}{0.95, 0.95, 0.92}
\newtcolorbox{promptbox}[2][]{
  enhanced, breakable, center title,
  left*=0pt, right*=0pt,
  boxsep=2pt, left=5pt, right=5pt,
  skin first=enhanced, skin middle=enhanced, skin last=enhanced,
  colback=backcolour,
  fonttitle=\bfseries\rmfamily,
  fontupper=\scriptsize,
  title={\footnotesize\strut{#2}},
  #1
}

\newtcolorbox{onebox}[2][]{
  enhanced, center title,
  left*=0pt, right*=0pt,
  boxsep=2pt, left=5pt, right=5pt,
  skin first=enhanced, skin middle=enhanced, skin last=enhanced,
  colframe=mybluetitle!90,
  colback=mybluetitle!10,
  fonttitle=\bfseries\rmfamily\fontfamily{phv}\selectfont,
  title={\footnotesize\strut{#2} \refstepcounter{subsubsection}%
    \addcontentsline{toc}{subsubsection}{\string\numberline{\thesubsubsection}#2}},
  #1
}

\definecolor{codegreen}{HTML}{478058}
\definecolor{codegray}{rgb}{0.5, 0.5, 0.5}
\definecolor{codepurple}{HTML}{4F5E80}
\lstdefinestyle{mystyle}{
  backgroundcolor=\color{backcolour},
  commentstyle=\color{codegreen},
  keywordstyle=\color{magenta},
  numberstyle=\tiny\color{codegray},
  stringstyle=\color{codepurple},
  basicstyle=\ttfamily\scriptsize,
  breakatwhitespace=false,
  breaklines=true,
  captionpos=b,
  keepspaces=true,
  frame=none,
  numbersep=5pt,
  showspaces=false,
  showstringspaces=false,
  showtabs=false,
  tabsize=2
}

  {\begin{enumerate}[leftmargin=4mm, itemsep=0pt, topsep=3pt]}%
  {\end{enumerate}}

\newcommand{\tableheadercell}[1]{\ensuremath{\vcenter{\hbox{\shortstack{\strut#1\strut}}}}}

\title{MedCaseAgent: From Multimodal Clinical Evidence to Rare Disease Case Reports}

\author[1,2]{Hongkang~Chu}
\author[1,3]{Yifan~Shen}
\author[4]{Zhuoqing~Zhong}
\author[3]{Tianjiao~Yu}
\author[1]{Jianguo~Cao}
\author[1,5,6,\dagger]{Meihuan~Huang}
\author[1,\dagger]{Xu~Cao}

\affiliation[1]{PediaMed AI}
\affiliation[2]{University of Chinese Academy of Sciences}
\affiliation[3]{University of Illinois Urbana-Champaign}
\affiliation[4]{Carnegie Mellon University}
\affiliation[5]{Shenzhen Children's Hospital}
\affiliation[6]{The Hong Kong Polytechnic University}

\contribution[\dagger]{Corresponding author}

\abstract{
Although Large Language Models (LLMs) have demonstrated potential in clinical documentation, generating the rigorous, evidence-based reasoning required for unprecedented, atypical, and rare disease case reports remains a significant challenge. We present MedCaseAgent, an end-to-end agentic reasoning paradigm that bridges this gap by synthesizing publication-quality reports from unstructured clinical notes. MedCaseAgent leverages a hierarchical structure, integrating patient-level data mining with iterative, tool-integrated reasoning. By dynamically querying medical literature and knowledge bases, the agent maintains high clinical fidelity and assesses case novelty. The framework is inherently modular, supporting various backbone LLMs in a plug-and-play fashion. To facilitate this research, we release MedCase-150K, a large-scale multimodal clinical case-report corpus, and MedCase-Bench, a robust temporally held-out evaluation suite. We also introduce an open-source model post-trained from Qwen3.5-9B and optimized specifically for this task. Under both LLM-judged and rule-based evaluation, MedCaseAgent outperforms existing general-purpose and specialized medical LLMs and agents. Blinded expert reviews and ablations further support the system's clinical relevance and evidence grounding.
}

\metadata[Demo]{\url{https://medcaseagent.medlastexam.org}}
\metadata[Website]{\url{https://huggingface.co/spaces/PediaMedAI/MedCaseAgent}}
\metadata[Correspondence to]{\email{xucao@pediamed.ai}}

\begin{document}

\maketitle

\begin{center}
  \includegraphics[width=\textwidth]{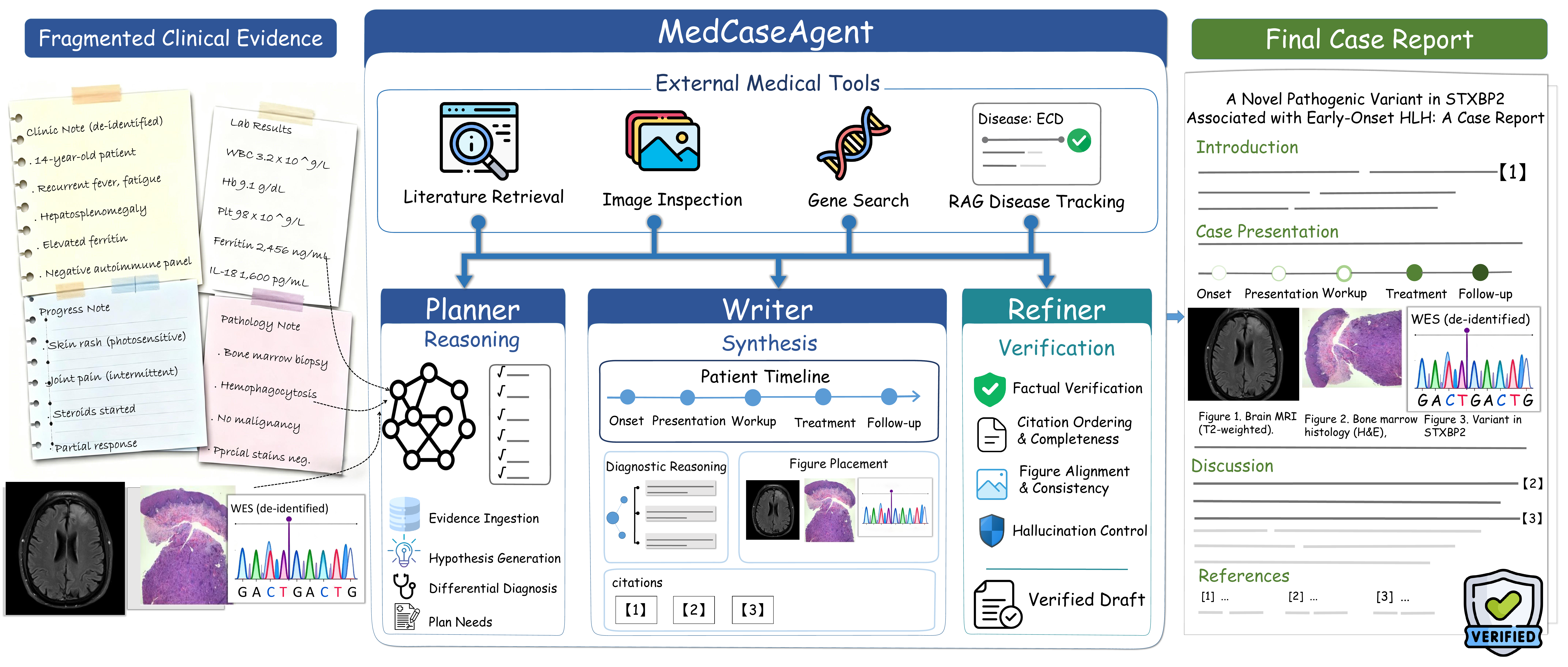}
\end{center}

\section{Introduction}
\label{sec:intro}

A clinical case report, often referred to as a case study, is a comprehensive and highly structured account of a patient's clinical encounter~\citep{cohen2006write,gagnier2014care,riley2017care}.  The defining characteristics of a compelling case report are its professionalism, novelty, and structural consistency. The presentation must be sufficiently rare, unique, or clinically significant to impart new medical knowledge. By publicly detailing these findings, case reports enable global medical communities to cross-reference phenotypically similar patients, accelerating the diagnosis of rare diseases, uncovering underlying molecular mechanisms, and facilitating larger-scale statistical analyses. Consequently, they serve as a vital cornerstone of medical progress, with nearly all top-tier medical journals dedicating specific categories to them and receiving a massive volume of global submissions. Despite their immense value, drafting these reports remains an arduous, entirely manual process for frontline physicians. Medical professionals must rigorously evaluate a case's novelty, meticulously synthesize patient histories and treatment plans from fragmented raw clinical notes, and conduct extensive literature reviews to identify homologous cases or genetic loci. Ultimately, they face the heavy cognitive burden of integrating specialty-specific reasoning with rigorous scientific citations to complete the narrative.

Recent years have witnessed significant advancements in large language models (LLMs), with systems such as GPT-5.4 and Gemini 3.1 Pro exhibiting remarkable reasoning capabilities. While their proficiency in structured problem-solving and logical inference has positioned them as the central cognitive engines for various clinical diagnostic agents~\citep{chiu2025simulating,wang2025medagent,zhao2026agentic}, these systems primarily focus on sequential diagnostic deduction rather than the generative synthesis of academic medical knowledge. Consequently, generating clinical case reports for rare diseases remains an exceptionally demanding frontier. It requires rigorous, multi-hop reasoning across highly fragmented and multimodal data streams, ranging from unstructured clinical shorthand and complex patient histories to rare imaging modalities and genomic sequences. Crucially, a viable case report generation pipeline must adhere to stringent factual constraints. It must strictly circumvent clinical hallucinations, such as spurious gene-literature matching, fabricated citations, or inaccurate patient summaries, which would entirely compromise the clinical validity of the report.

To address this gap, we propose MedCaseAgent, an end-to-end agentic framework that synthesizes academic-grade case reports from unstructured clinical notes. Structured as an interactive planner-writer-editor workflow, the agent orchestrates specialized tools for literature retrieval, exact citation tracking, image analysis, and genetic search to guarantee clinical accuracy and reduce hallucinations. In parallel, we construct MedCase-150K from over 7 million PubMed articles published since 2000 and derive MedCase-Bench as the recent held-out evaluation split. By automatically stratifying cases by clinical rarity, these resources provide a real-world simulation that tests how effectively models interpret fragmented multimodal evidence to formulate complete reports.

Our contributions can be summarized as follows:

\begin{enumerate}[label=(\arabic*), leftmargin=*]
    \item We propose \textbf{MedCaseAgent}, a plug-and-play clinical case report agent that integrates multimodal foundation models with dynamic tool-use mechanisms to actively orchestrate and synthesize complex clinical narratives from clinical notes.
    \item We introduce \textbf{MedCase-150K}, a large-scale collection of rare and novel disease clinical case reports, and \textbf{MedCase-Bench}, a temporally held-out subset derived from MedCase-150K, both designed for multimodal medical case report generation and rigorously stratified by clinical rarity.
    \item We validate MedCaseAgent comprehensively on MedCase-Bench, where it surpasses mainstream MLLMs and medical agentic systems. Clinician evaluations and a widespread user study further highlight the reliability and usefulness of MedCaseAgent.
\end{enumerate}

\FloatBarrier
\section{Related Work}

\noindent\textbf{Agentic Systems in Clinical Practice.} Recent clinical agent research first shows strong progress in imaging. Across radiology and pathology, agentic systems support image-grounded reasoning~\citep{du2026care,qiu2025quantifying,chen2025pathagent}, evidence retrieval~\citep{cao2026atpo}, report generation~\citep{qi2026eviagent,zhong2025vision,wang2026qcagent}, and quality control, but most remain designed for classification~\citep{fard2025ft}, segmentation~\citep{liu2026medsam}, question answering~\citep{he2025astrid,li2024mediq,xiao2025dynamic,xu2025lingshu,jiang2025hulu}, or planning~\citep{zhang2025csap}. Text-centered agents then extend these ideas to longitudinal records~\citep{li2025care}, simulated hospital workflows~\citep{li2024agent}, clinical communication~\citep{zhi2025reinventing}, and safety oversight~\citep{kim2025tiered}. Systems such as AgentEHR~\citep{liao2026agentehr}, MedAgentGym~\citep{xu2025medagentgym}, and Reflectool~\citep{liao2025reflectool} study streaming EHRs~\citep{croxford2025evaluating}, retrospective summaries, and structured clinical tasks~\citep{ren2025healthcare,zhong2025multi}. A related direction studies clinical diagnosis and agent architecture search with agentic reasoning systems~\citep{ghezloo2025pathfinder,zhao2026agentic,xia2025mmedagent,pushkar2025teammedagents,yang2026lungnoduleagent,liu2025dragent,wu2025lessons,zhuang2025learning}. These methods improve traceability through retrieval~\citep{wang2026qcagent,deng2026multidx,li2026medcorag,rose2025meddxagent,ahmed2026provenance}, verification~\citep{chen2025pathagent,wang2026qcagent,li2026medcorag,ahmed2026provenance}, and iterative refinement~\citep{chen2025pathagent,wang2026qcagent,li2026medcorag,rose2025meddxagent}, but usually stop at prediction~\citep{xia2025mmedagent,pushkar2025teammedagents,deng2026multidx,yang2026lungnoduleagent,liu2025dragent}, recommendation~\citep{wu2025lessons}, or short-form clinical writing~\citep{wang2026qcagent}. By contrast, MedCaseAgent targets end-to-end case report synthesis, linking multimodal evidence, timeline reconstruction, literature support, and manuscript-level writing.

\noindent\textbf{Automated Scientific and Case-Report Writing.} General-purpose research-writing agents such as the AI Scientist~\citep{lu2024ai,yamada2025ai}, data-to-paper~\citep{ifargan2024autonomous}, CycleResearcher~\citep{weng2024cycleresearcher}, and PaperOrchestra~\citep{song2026paperorchestra} automate machine-learning or data-driven papers, but do not address patient-grounded clinical evidence, rarity assessment, or medical citation grounding. In medicine, prior studies prompted ChatGPT to draft individual case reports and concluded that the outputs required substantial expert revision and could contain fabricated references~\citep{buholayka2023readiness,pinto2024comparison}; case-report corpora have otherwise been used for diagnostic reasoning~\citep{wu2025medcasereasoning} or information extraction~\citep{zhang2025casereportbench}. MedCaseAgent instead treats case-report writing as a trainable, tool-grounded agentic task with a dedicated benchmark.
\FloatBarrier

\raggedbottom
\setcounter{topnumber}{1}
\section{Methodology}
\label{sec:methodology}

\begin{figure}[!t]
  \centering
  \includegraphics[width=0.90\linewidth]{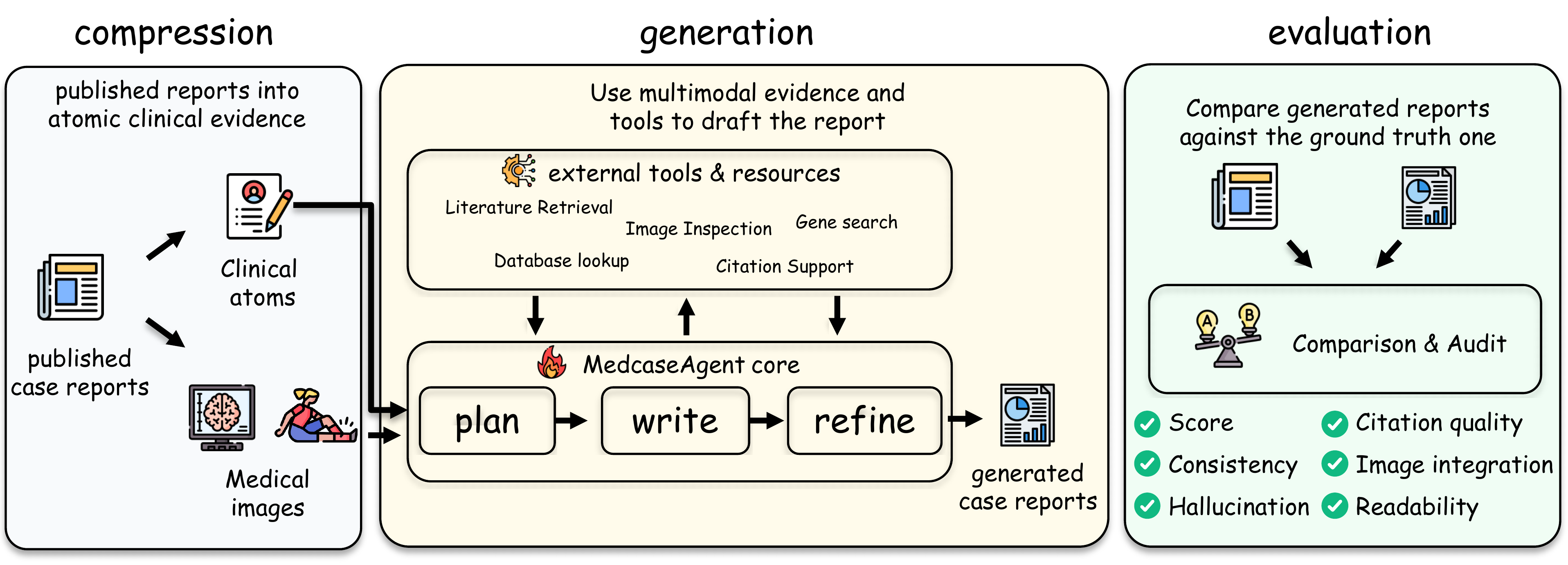}
  \caption{Overview of the MedCaseAgent report-generation workflow.}
  \label{fig:medcaseagent_framework_overview}
  \vspace{-3mm}
\end{figure}

\paragraph{Overview} We first construct MedCase-150K, a large-scale collection of peer-reviewed clinical case reports. From this collection, we define MedCase-Bench as a temporally held-out evaluation subset of 584 case reports published after January 2026. MedCase-Bench will be open-sourced under CC-BY-NC 4.0. We then propose MedCaseAgent (Figure~\ref{fig:medcaseagent_framework_overview}), an agentic report-writing system that plans over the fragmented evidence, retrieves supporting medical literature, drafts the manuscript, and audits the final report.

\subsection{Task Formulation}
\begin{figure}[t]
  \centering
  \includegraphics[width=0.95\linewidth]{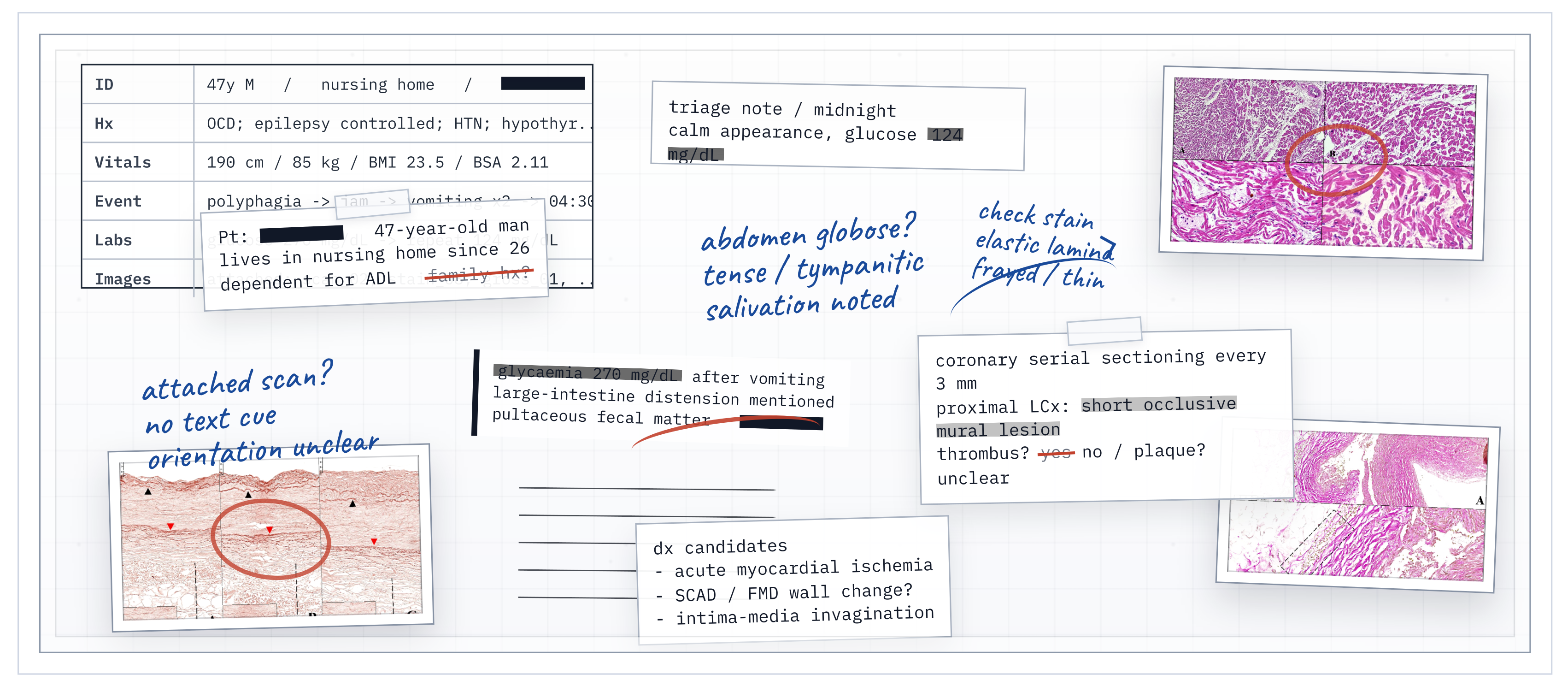}
  \caption{Illustration of fragmented first-hand clinical observations used as multimodal input to MedCaseAgent. The notes are patient-specific and incomplete, resembling the partial facts available before a formal academic case report is written.}
  \label{fig:messy_notes_input}
  \vspace{-3mm}
\end{figure}

Given unstructured clinical observations $X$ and external medical resources $\mathcal{D}$, the goal is to generate a structured clinical case report $Y_{\text{report}}$. The inputs include textual clinical notes $X_{\text{notes}} = \{x_1, x_2, \dots, x_n\}$ that describe patient demographics, symptoms, laboratory findings, diagnosis, treatment, and outcome, together with diagnostic images $X_{\text{img}} = \{i_1, i_2, \dots, i_m\}$ such as MRI, CT, radiography, or histology scans. We represent the textual input as clinical atoms grouped into history, presentation, diagnostics, management, and outcome. This atom format is designed to approximate the first-hand information that clinicians receive before a manuscript is written: partial notes, test results, image findings, missed visits, and follow-up details, rather than a polished academic narrative~\citep{abacha2023empirical}. Recent work also suggests that LLM-written summaries can form portable representations for irregular clinical time series~\citep{ji2026can}; in our setting, atoms are used as controllable evidence for manuscript generation rather than as predictive embeddings. The atom format also gives a stable supervision unit for planner training and reward computation, since each generated plan can be checked for atom coverage. The target output is a complete report with source-driven manuscript sections such as Introduction, Case Description, Discussion, and References. During generation, the vision-language model acts as an agent. It issues tool calls and search queries $q_t$ to retrieve supporting knowledge $Z_{\text{retrieved}}$ from $\mathcal{D}$, and then links the raw clinical atoms with a complete manuscript. Figure~\ref{fig:messy_notes_input} illustrates fragmented and partially structured clinical observations as the multimodal input for MedCaseAgent; Appendix~\ref{subsec:medcaseagent_workflow_details} provides a fuller prompt-output atom-compression view.

\subsection{MedCase-150K and MedCase-Bench Construction}
While datasets like CaseReportBench~\citep{zhang2025casereportbench} treat case reports as targets for information extraction, our setting requires a model to write a complete, well-cited report from fragmented multimodal inputs, which is a much harder problem. As there is no existing dataset for this domain, we construct MedCase-150K, a dataset of 154,872 case reports describing rare diseases and unprecedented or atypical presentations (see Appendix~\ref{subsec:medcase150k_data_construction_views}, Data Construction and Validation, for dataset overview and statistics). Each MedCase-150K entry contains a published source article, unpaired medical images, and the corresponding LLM-compressed clinical atoms. The construction pipeline of MedCase-150K is semi-automated. First, we manually identify PMC Open Access case-report publications with full-text XML and high-quality medical images. An LLM-based parser converts the final prose into unstructured clinical atoms $X_{\text{notes}}$ by removing background literature, reference-driven claims, and final diagnostic exposition. The parser keeps patient-specific facts, including timeline details, tests, treatments, image findings, and outcomes.

This compression step is necessary because the original article already contains the target report. By turning the published manuscript back into noisy clinical atoms, we make the task closer to real clinical writing and create supervision that can be used for planner SFT and GRPO training. Source images accompany the atoms without captions or explicit image--text links. Human reviewers checked all 632 benchmark candidates for atom fidelity, then screened retained cases for rarity-label validity, retaining 584 cases (Appendix~\ref{subsec:atom_human_verification}). MedCase-Bench is derived from MedCase-150K as the evaluation subset: it uses a strict temporal holdout of peer-reviewed PMC Open Access case reports published after January 2026, reducing the risk that evaluated models have memorized the final reports.

\subsection{MedCaseAgent: Agentic Medical Case Report Synthesis Pipeline}
Given fragmented clinical atoms and paired medical images, MedCaseAgent divides report generation into three stages: Planner, Writer, and Editor. The same external tools are available throughout all three stages. In practice, tool use is intentionally concentrated in the Planner, where most literature search, image inspection, gene lookup, and prior-case retrieval calls are made before manuscript drafting begins.

\textbf{Phase 1: The Planner (Multimodal Reasoning and External Tool Execution).} 
The planner starts from the multimodal state $S_0 = (X_{\text{notes}}, X_{\text{img}})$. A multimodal vision-language model reads the images and clinical atoms, then selects actions from a predefined action space~\citep{yao2022react, akkiraju2024facts}, following prior tool-mediated clinical agent systems~\citep{yue2024clinicalagent}.
These tools include Literature Retrieval for PubMed search and citation grounding, Image Inspection for single-frame or multi-panel figure interpretation, Gene Search for ClinGen variant lookup, and Disease-Index RAG for retrieving similar MedCase-150K prior cases to support novelty planning. The Disease-Index RAG corpus excludes the evaluated cases' source articles and related literature to prevent information leakage.
At step $t$, the planner generates a query $q_t$, retrieves context $C_t$, and updates working memory as $S_{t+1} = S_t \cup C_t$.

\begin{figure}[t]
  \centering
  \includegraphics[width=0.95\linewidth]{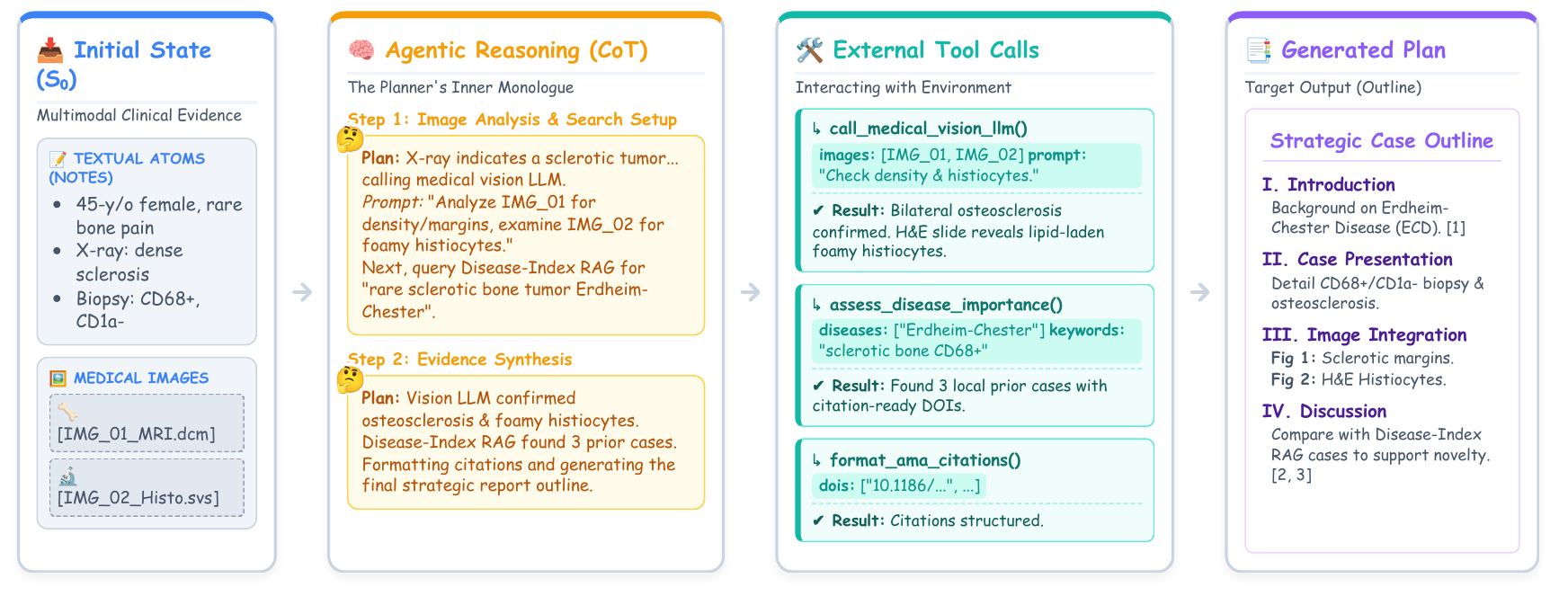}
  \caption{Planner phase from initial multimodal evidence to a generated case-report plan.}
  \label{fig:phase1}
  \vspace{-3mm}
\end{figure}

The planner returns three outputs. First, it maps each clinical atom to the source-driven manuscript section in which it should appear, so that patient history, diagnostic tests, procedures, outcomes, and unresolved details are not lost. Second, it builds an evidence plan by selecting literature search terms, verified citations, and Disease-Index RAG results that can support background and discussion claims. Third, it creates a figure plan by assigning each source image to a main-text callout and a short caption. This plan gives the writer a grounded intermediate representation between raw notes and manuscript structure. Figure~\ref{fig:phase1} illustrates how the planner moves from the initial multimodal state through reasoning and tool calls to the generated section, citation, and figure plan.

\textbf{Phase 2: The Writer (Draft Generation).} 
The writer creates the first manuscript draft from the planner output and the accumulated evidence $Z_{\text{retrieved}} = \bigcup_{t=1}^{T} C_t$. It conditions on the clinical notes, images, and retrieved context:
\begin{equation}
    P(Y_{\text{draft}} \mid X_{\text{notes}}, X_{\text{img}}, Z_{\text{retrieved}})
\end{equation}
The writer follows the planner section by section, which helps keep manuscript-length generation coherent in a setting where long-output generation remains challenging for many long-context LLMs~\citep{bai2024longwriter}. It writes patient-specific content only from the atoms and uses retrieved literature only for background, diagnostic criteria, treatment rationale, and comparison with prior work. This separation is important because patient facts should not be cited as if they came from the literature, and literature facts should not be copied into the case description as patient evidence.

\textbf{Phase 3: The Editor (Refinement and Verification).} 
The editor refines the draft and checks factual consistency. It creates required front matter such as the title and abstract, verifies that the final text $Y_{\text{report}}$ is supported by the clinical atoms $x_i \in X_{\text{notes}}$, checks citation order and density, and confirms that every figure is described in the main text. The editor also checks that the draft follows the source section order, that broad case-report sections still contain CARE-style content, and that unsupported statements are removed or rewritten. In the implemented pipeline, later audit and citation-repair passes reuse the same evidence bank to correct missing references, citation order, and figure callouts. The final output is a verified manuscript.

\subsection{Training Objective for the Planner}
\label{subsec:training_objective}

\begin{wrapfigure}{r}{0.6\textwidth}
  \vspace{-0.85\baselineskip}
  \centering
  \includegraphics[width=\linewidth]{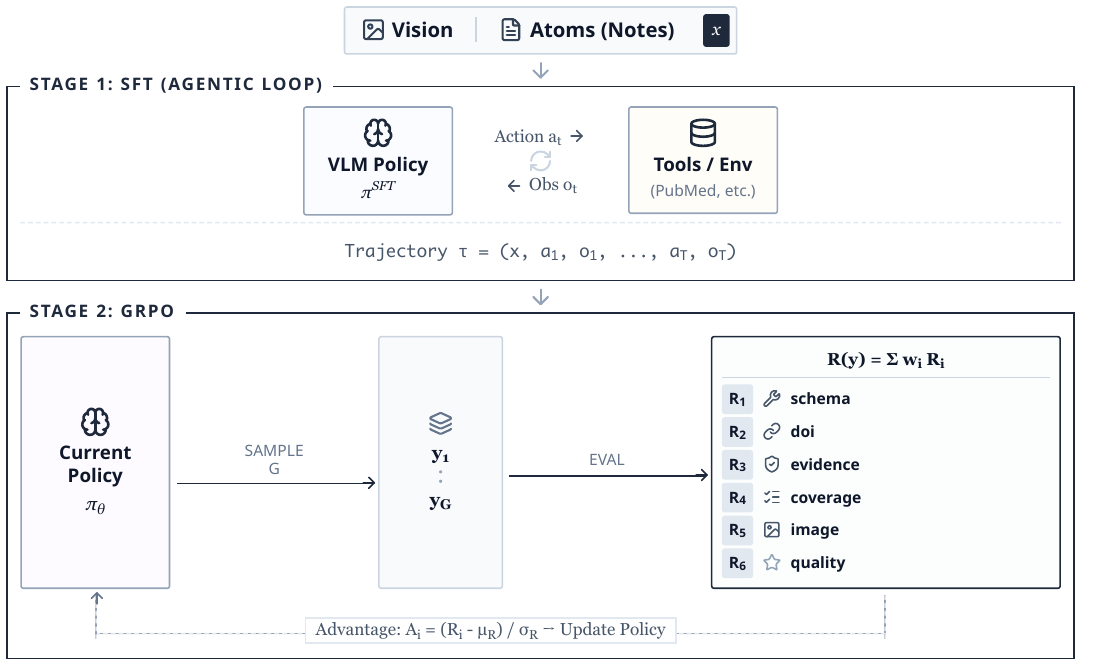}
  \caption{Training phase for the MedCaseAgent planner.}
  \label{fig:training}
  \vspace{-0.6\baselineskip}
\end{wrapfigure}

We train the phase 1 planner with a two-stage recipe implemented with SWIFT and LoRA adapters. Figure~\ref{fig:training} summarizes the planner-training flow from multimodal inputs through SFT trajectories and GRPO reward feedback. This use of supervised trajectories followed by verifiable reward feedback follows the GRPO-style reasoning-training line~\citep{shao2024deepseekmath} and is aligned with recent medical reasoning work on verifiable problems, process supervision, and reinforcement learning~\citep{chen2025towards,jiang2026meds3,liu2025fleming}. First, we apply supervised fine-tuning (SFT) on teacher-generated planner trajectories. The data preparation converts the public phase 1 planner trajectory into agent messages, with tool calls, tool responses, and final plans used as supervision. Given an input case $x$ and a teacher trajectory $\tau^{*}=(y_1^{*},\ldots,y_T^{*})$, the SFT loss is $\mathcal{L}_{\mathrm{SFT}}=-\sum_{t=1}^{T}\log \pi_{\theta}(y_t^{*}\mid x,y_{<t}^{*})$.

Second, we apply Group Relative Policy Optimization (GRPO) to improve tool-use decisions and plan quality. For each case, the current policy samples a group of planner trajectories $\{\tau_i\}_{i=1}^{G}$. Each trajectory is scored by a compact reward:
\begin{equation}
\begin{aligned}
    R(\tau)
    ={}&
    w_{\mathrm{schema}}R_{\mathrm{schema}}
    +
    w_{\mathrm{doi}}R_{\mathrm{doi}}
    +
    w_{\mathrm{evidence}}R_{\mathrm{evidence}}
    \\
    &+
    w_{\mathrm{coverage}}R_{\mathrm{coverage}}
    +
    w_{\mathrm{image}}R_{\mathrm{image}}
    +
    w_{\mathrm{quality}}R_{\mathrm{quality}}.
\end{aligned}
\end{equation}

\Needspace{0.36\textheight}

Here, $R_{\mathrm{schema}}$ checks allowed tool names and valid arguments; $R_{\mathrm{doi}}$ checks whether the generated search queries retrieve the scorer-only source DOI when available; $R_{\mathrm{evidence}}$ checks DOI validity, citation anchors, and overlap with non-source retrieved support; $R_{\mathrm{coverage}}$ checks clinical atom coverage and required section coverage; $R_{\mathrm{image}}$ checks structure and image placement; and $R_{\mathrm{quality}}$ uses a deterministic quality proxy or a stored judge score. $R_{\mathrm{doi}}$ provides a dense signal for query quality: recovering the one article that describes the same patient requires queries that capture the distinctive clinical features of the case, the same capability that retrieving analogous prior cases demands.

The group-relative advantage is computed as $A_i=(R(\tau_i)-\mu_R)/(\sigma_R+\epsilon)$, where $\mu_R$ and $\sigma_R$ are the mean and standard deviation of rewards within the sampled group. The default weights prioritize clinical content and evidence: DOI retrieval, evidence support, coverage, and quality receive 0.20 each, while schema compliance and structure/image placement receive 0.10 each. Appendix~\ref{subsec:grpo_reward_sensitivity} provides the rationale and reward-weight sensitivity analysis.

\FloatBarrier
\setcounter{topnumber}{2}
\flushbottom

\section{Experiments}
\label{sec:experiments}

\subsection{Experimental Setup}

\textbf{Baselines.} For benchmark evaluation, all systems are evaluated on MedCase-Bench. The input is the patient atom notes and unlabeled medical images. For a fair comparison, all baseline pipelines are equipped with the same tools. We reserve standalone model names for direct tool-equipped LLM/VLM baselines; in backbone-swap experiments, \emph{MedCaseAgent (model)} denotes the full MedCaseAgent pipeline using that model as the centered LLM. MedCaseAgent-Base uses Qwen3.5-9B as its backbone. MedCaseAgent-Tuned uses the same Qwen3.5-9B backbone with the planner trained using the objective in Section~\ref{subsec:training_objective}; implementation and compute details are provided in Appendix~\ref{subsec:compute_resources}. MedCaseAgent-Max uses GPT-5.4.

\begin{wrapfigure}{R}{0.65\textwidth}
    \vspace{-0.75\baselineskip}
    \centering
    \includegraphics[width=\linewidth]{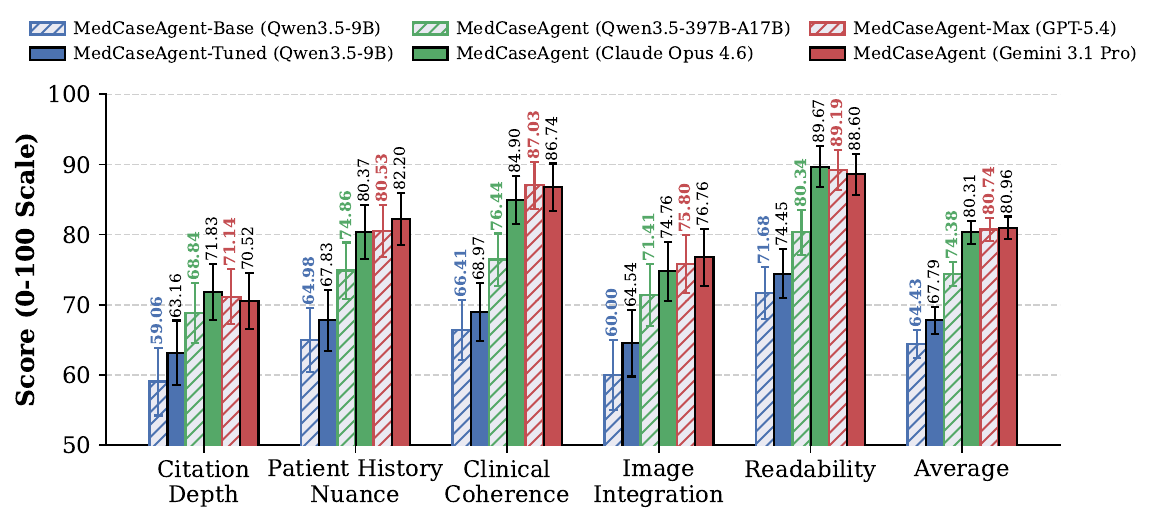}
    \caption{MedCaseAgent configuration comparison on MedCase-Bench across five rubric metrics and their average. Each legend entry is a MedCaseAgent variant, with the centered LLM shown in parentheses. Bars show means; error bars show standard deviations.}
    \label{fig:llm_configuration_scores}
    \vspace{-0.8\baselineskip}
\end{wrapfigure}

Direct baselines include tool-equipped general VLMs such as Gemma-4-31B~\citep{gemma4modelcard2026} and specialist medical VLMs such as the MedGemma family~\citep{sellergren2025medgemma,sellergren2026medgemma}. We compare with representative medical agentic systems, including MedAgents~\citep{tang2024medagents}, MDAgents~\citep{kim2024mdagents}, MedAgent-Pro~\citep{wang2025medagent}, and MMedAgent~\citep{li2024mmedagent}. These methods are strong reference points because they use agents to support clinical reasoning, diagnosis, and medical decision-making. We adapt them to the same report-generation setting and provide the same tool access, so the comparison focuses on system design rather than tool availability.

\textbf{Judge ensemble and report scoring.} For report scoring, we use a four-model VLM judge ensemble (Qwen3.5-397B-A17B, Gemini 3.1 Pro, Claude Opus 4.6, and GPT-5.4), following clinical LLM-as-judge and automated medical-reasoning evaluation practice~\citep{croxford2025evaluating,zhou2025automating}. The ensemble spans four model families, so no single family determines the score. Appendix~\ref{subsec:llm_clinician_agreement} reports agreement with clinical experts, and Appendix~\ref{subsec:judge_robustness} reports per-judge scores.

\textbf{Evaluation metrics.} We evaluate generated reports from complementary views. Rubric-level judge scores assess citation depth, patient-history nuance, clinical coherence, image integration, and readability on a 0--10 scale; we linearly scale them to 0--100 for display and report average quality as the mean of these five dimensions. Disease-subset scores test robustness across common, rare, and ultra-rare diagnoses; rule-based metrics audit retrieval, novelty recognition, figure ordering, and citation coverage; and clinician-reviewer A/B validation measures practical preference and authorship indistinguishability.

\subsection{Report Quality and Disease-Stratified Evaluation}
\label{subsec:overall_results}

Figure~\ref{fig:llm_configuration_scores} compares MedCaseAgent configurations on MedCase-Bench when the same agent pipeline is equipped with different backbone LLMs. After training, MedCaseAgent-Tuned improves over MedCaseAgent-Base in average score ($64.43\!\to\!67.79$), with clear gains in Citation Depth ($59.06\!\to\!63.16$) and Image Integration ($60.00\!\to\!64.54$). Stronger backbones further raise overall report quality within the MedCaseAgent workflow, with MedCaseAgent-Max (GPT-5.4), MedCaseAgent (Gemini 3.1 Pro), and MedCaseAgent (Claude Opus 4.6) forming the top group.

\begin{table*}[!t]
    \centering
    \caption{Disease-subset evaluation on MedCase-150K, comparing tool-equipped general MLLMs and medical agentic systems across selected disease categories. Each disease column samples 100 cases, except Erdheim-Chester disease (ECD), where all 53 available MedCase-150K cases are used. Each score averages citation depth, patient-history nuance, clinical coherence, image integration, and readability.}
    \label{tab:disease_category_scores}
    \setlength{\tabcolsep}{3.5pt}
    \renewcommand\arraystretch{1.12}
    \rowcolors{2}{gray!10}{white}
    \resizebox{1.0\textwidth}{!}{%
      \begin{tabular}{llcccccc}
        \toprule
        \rowcolor{gray!25}
        \textbf{Method} & \textbf{Centered LLM} & \textbf{COVID-19} & \tableheadercell{\textbf{Small-Cell}\\\textbf{Lung Cancer}} & \tableheadercell{\textbf{Erdheim-Chester}\\\textbf{Disease}} & \tableheadercell{\textbf{Rare Genetic}\\\textbf{Disorder}} & \tableheadercell{\textbf{Ultra-Rare}\\\textbf{Disease}} & \textbf{Average} \\
        \midrule
        Human Publication & - & 82.96 & 76.84 & 73.75 & 73.28 & 71.07 & 75.58 \\
        \midrule
        Qwen3.5-9B~\citep{yang2025qwen3} & - & 62.46 & 55.84 & 54.35 & 53.44 & 53.79 & 55.98 \\
        Qwen3.5-397B-A17B & - & 78.28 & 74.64 & 66.21 & 66.15 & 64.11 & 69.88 \\
        Gemma-4-31B & - & 72.34 & 69.26 & 64.25 & 63.03 & 62.45 & 66.27 \\
        Claude Opus 4.6 & - & 82.05 & 75.18 & 74.62 & 73.58 & 67.75 & 74.64 \\
        GPT-5.4 & - & 84.41 & 79.41 & 74.34 & 70.65 & 73.36 & 76.43 \\
        Gemini 3.1 Pro & - & 82.42 & 75.92 & 72.30 & 75.61 & 72.09 & 75.67 \\
        \midrule
        LLaVA-Med-7B~\citep{li2023llava} & Mistral-7B & 52.85 & 49.27 & 46.68 & 47.45 & 43.90 & 48.03 \\
        MedGemma-1.5-4B~\citep{sellergren2026medgemma} & Gemma 3-4B & 51.61 & 49.16 & 46.26 & 47.57 & 40.41 & 47.00 \\
        MedVLM-R1-2B~\citep{pan2025medvlm} & Qwen2-VL-2B & 46.88 & 40.84 & 39.50 & 38.08 & 34.31 & 39.92 \\
        \midrule
        MedAgents~\citep{tang2024medagents} & Qwen3.5-9B & 66.37 & 60.03 & 58.00 & 57.37 & 55.64 & 59.48 \\
        MDAgents~\citep{kim2024mdagents} & Qwen3.5-9B & 66.22 & 61.80 & 59.18 & 58.40 & 56.93 & 60.51 \\
        MedAgent-Pro~\citep{wang2025medagent} & Qwen3.5-9B & 70.58 & 64.52 & 62.61 & 61.68 & 58.16 & 63.51 \\
        MMedAgent~\citep{li2024mmedagent} & Qwen3.5-9B & 69.52 & 63.93 & 62.34 & 61.70 & 58.71 & 63.24 \\
        \midrule
        MedCaseAgent-Base & Qwen3.5-9B & 72.88 & 66.40 & 64.23 & 63.43 & 58.95 & 65.18 \\
        MedCaseAgent-Max & GPT-5.4 & 87.98 & 82.97 & 82.80 & 83.71 & 78.37 & 83.17 \\
        \bottomrule
      \end{tabular}
    }
    \vspace{-2mm}
\end{table*}

Table~\ref{tab:disease_category_scores} stratifies performance across disease-specific subsets sampled from MedCase-150K and tests whether models remain stable as disease priors weaken. Although scores tend to decline from better-represented diseases such as COVID-19 and small-cell lung cancer to rarer settings such as Erdheim-Chester disease, rare genetic disorders, and ultra-rare diseases, MedCaseAgent-Max remains the best-performing system across all disease subsets. Frontier closed-source LLMs score close to the human-publication reference on average, and MedCaseAgent-Max scores above it on every disease subset. In the blinded study, 26 of 32 experienced clinicians correctly identified the generated report in at most three of five pairs, and reviewers more often preferred the MedCaseAgent-Max report (Appendix~\ref{subsec:clinician_blinded_validation}). MedCaseAgent-Base also stays above prior medical-agent baselines, indicating that the planner--writer--editor workflow improves disease-level robustness even before the strongest full configuration is used. Appendix~\ref{subsec:llm_judged_ab_appendix} reports the corresponding LLM-judged blind A/B comparison across baselines, and Appendix~\ref{subsec:generated_case_report_sheets} provides three complete case reports generated by MedCaseAgent for qualitative inspection.

\subsection{Rule-Based Structural Metrics and Ablation}
\label{subsec:rule_based_metrics}

\begin{table*}[!t]
    \centering
    \caption{Comparison of rule-based and structural evaluation metrics across different models. All entries are percentages. Abbreviations: Exact DOI Retrieval (Exact DOI), Novelty Recognition (Novel Rec.), Figure Sequence Accuracy (Fig. Seq.), Authorship Indistinguishability (Auth. Indist.), and citation recall for must-cite (P0) and good-to-cite (P1) references.}
    \label{tab:rule_based_metrics}
    \setlength{\tabcolsep}{3.5pt}
    \renewcommand\arraystretch{1.08}
    \rowcolors{3}{gray!10}{white}
    \resizebox{0.80\textwidth}{!}{%
      \begin{tabular}{lcccccc}
        \toprule
        \rowcolor{gray!25}
        \textbf{Method} & \textbf{Exact DOI} & \textbf{Novel Rec.} & \textbf{Fig. Seq.} & \textbf{Auth. Indist.} & \textbf{P0 Recall} & \textbf{P1 Recall} \\
        \midrule
        Qwen3.5-9B & 22.00 & 38.12 & 54.75 & 31.95 & 62.63 & 31.54 \\
        Qwen3.5-397B-A17B & 36.27 & 57.39 & 66.93 & 45.53 & 66.70 & 35.32 \\
        Gemma-4-31B & 31.11 & 52.88 & 63.36 & 38.92 & 65.87 & 34.87 \\
        Claude Opus 4.6 & 43.22 & 66.75 & 73.45 & 53.79 & 68.60 & 35.62 \\
        GPT-5.4 & 46.90 & 69.01 & 75.92 & 50.89 & 69.15 & 36.13 \\
        Gemini 3.1 Pro & 44.50 & 68.01 & 76.56 & 52.76 & 68.83 & 35.59 \\
        \midrule
        MedCaseAgent-Base & 30.60 & 52.40 & 61.10 & 42.84 & 64.62 & 34.16 \\
        MedCaseAgent-Max & 84.59 & 88.35 & 87.34 & 69.18 & 74.14 & 44.82 \\
        \bottomrule
      \end{tabular}
    }
    \vspace{-2mm}
\end{table*}

Table~\ref{tab:rule_based_metrics} reports rule-based and structural metrics for source retrieval, novelty recognition, figure ordering, authorship indistinguishability, and citation coverage. Exact DOI measures whether a keyword-based literature search retrieves the ground-truth PubMed source paper from which the atoms were derived. Note that even if the ground-truth paper is retrieved, it is masked during report generation to prevent information leakage. Exact DOI is thus a per-case source-retrieval hit rate: it reflects whether the agent's queries are specific enough to surface the single article describing the same patient among millions of PubMed records. Novelty Recognition is the percentage of cases where the model recognizes rare or first-case novelty even though this cue is omitted from the atoms. Figure Sequence Accuracy is binary per case: 1 only when the generated report uses the same figure order as the published reference, and 0 otherwise. Authorship Indistinguishability follows an LLM-based A/B test and measures how often generated reports are difficult to distinguish from the published ones.
Inspired by PaperOrchestra~\citep{song2026paperorchestra}, we assess citation coverage using two reference tiers. P0 denotes must-cite analogous or claim-critical case-report sources. P1 denotes useful background, mechanism, treatment, or comparison sources.
After canonical PubMed/DOI/title matching, tier $k$ is scored as $\mathrm{Recall}_k=|C\cap G_k|/|G_k|$, where $C$ is the generated citation set and $G_k$ is the reference set for that tier. MedCaseAgent-Max has the strongest structural grounding (84.59\% Exact DOI vs.\ 46.90\% for the strongest direct baseline), the best novelty recognition, the closest figure sequencing, and the highest P0/P1 Recall.

\label{subsec:tool_ablation}

\begin{table*}[!t]
    \centering
    \caption{Two ablation studies for planner training (upper part) and external tools (lower part) using report-quality metrics and novelty recognition. Novelty Recognition is the same percentage metric as Novel Rec.\ in Table~\ref{tab:rule_based_metrics}. The lower part is performed on MedCaseAgent-Max.}
    \label{tab:tool_ablation}
    \setlength{\tabcolsep}{2.4pt}
    \renewcommand\arraystretch{1.08}
    \rowcolors{2}{gray!10}{white}
      \begin{adjustbox}{max width=0.90\linewidth}
\begin{tabular}{lccccccc}
        \toprule
        \rowcolor{gray!25}
        \textbf{Configuration} & \tableheadercell{\textbf{Citation}\\\textbf{Depth}} & \tableheadercell{\textbf{Patient-History}\\\textbf{Nuance}} & \tableheadercell{\textbf{Clinical}\\\textbf{Coherence}} & \tableheadercell{\textbf{Novelty}\\\textbf{Recognition}} & \tableheadercell{\textbf{Image}\\\textbf{Integration}} & \textbf{Readability} & \textbf{Average} \\
        \midrule
        MedCaseAgent-Base & 59.06 & 64.98 & 66.41 & 52.40 & 60.00 & 71.68 & 62.42 \\
        MedCaseAgent-Tuned & 63.16 & 67.83 & 68.97 & 59.80 & 64.54 & 74.45 & 66.46 \\
        \midrule
        MedCaseAgent-Max & \textbf{71.14} & \textbf{80.53} & \textbf{87.03} & \textbf{88.35} & \textbf{75.80} & \textbf{89.19} & \textbf{82.01} \\
        \quad w/o Literature Retrieval & 48.72 & 78.64 & 79.18 & 72.48 & 75.12 & 87.03 & 73.53 \\
        \quad w/o Image Inspection & 70.28 & 78.91 & 84.10 & 85.71 & 58.95 & 86.85 & 77.47 \\
        \quad w/o Gene Search & 67.92 & 79.64 & 84.86 & 81.69 & 75.14 & 88.10 & 79.56 \\
        \quad w/o Disease-Index RAG & 62.84 & 78.73 & 81.48 & 68.90 & 74.92 & 87.30 & 75.70 \\
        \bottomrule
      \end{tabular}
\end{adjustbox}
\vspace{-3mm}
\end{table*}

Table~\ref{tab:tool_ablation} uses the same five report-quality dimensions as Figure~\ref{fig:llm_configuration_scores} to study planner training and tool removal, and it adds Novelty Recognition, the same percentage metric abbreviated as Novel Rec.\ in Table~\ref{tab:rule_based_metrics}, as the structural metric most directly tied to rarity and prior-case lookup. The first three rows isolate the effect of replacing the untuned Qwen3.5-9B planner with the trained planner and then with the strongest full system. Among tool ablations, removing literature retrieval causes the largest drop in average score and citation depth. Image inspection and disease-index retrieval contribute most to image integration and novelty recognition, respectively.

\paragraph{Reward-weight sensitivity}
We compare four GRPO reward-weight profiles retaining the source-DOI reward from the same SFT checkpoint under fixed training and evaluation settings. The default achieves the highest four-metric average (71.68), with a 1.03-point spread across these profiles (Appendix~\ref{subsec:grpo_reward_sensitivity}, Table~\ref{tab:grpo_reward_sensitivity}). Removing the source-DOI reward ($w_{\mathrm{doi}}=0$, remaining weights renormalized) lowers the average to 69.83, a decrease of 1.85 points, with novelty recognition declining from 59.08\% to 55.78\% and report quality from 67.23 to 65.37.

\subsection{Clinical Evaluation and User Study}
\label{subsec:clinical_evaluation_user_study}

\begin{wraptable}{r}{0.42\textwidth}
    \vspace{-\baselineskip}
    \centering
    \captionsetup{font=footnotesize,skip=2pt}
    \caption{Mean expert scores (1--10; $\uparrow$).}
    \label{tab:hospital_expert_evaluation}
    \setlength{\tabcolsep}{4pt}
    \renewcommand\arraystretch{1.0}
    \rowcolors{2}{gray!10}{white}
    \begin{tabular}{lccc}
        \toprule
        \rowcolor{gray!25}
        \textbf{Report} & \textbf{Quality} & \textbf{Timeline} & \tableheadercell{\textbf{Omission}\\\textbf{control}} \\
        \midrule
        Raw records & 7.4 & 7.5 & 7.8 \\
        Atoms & 7.6 & 7.6 & 7.6 \\
        Clinician & 7.5 & 7.8 & 7.9 \\
        \bottomrule
    \end{tabular}
    \vspace{-0.65\baselineskip}
\end{wraptable}

To assess whether MedCaseAgent can support case-report writing from real clinical data, we conducted a hospital-based study using 30 de-identified multimodal cases from 10 hospitals. All cases were selected as suitable for rare-disease case-report preparation, including cases with unpublished manuscripts intended for submission. Fourteen clinical experts assessed reports generated from raw records and report-derived atoms alongside clinician-written reports; each case was independently reviewed by three experts against the source records. Table~\ref{tab:hospital_expert_evaluation} shows comparable overall quality for raw-record and atom-based generation (7.4 versus 7.6), with higher omission-control scores for raw input (7.8 versus 7.6). Appendix~\ref{subsec:hospital_evaluation_protocol} details the input differences. Because these cases have no published source article, they evaluate MedCaseAgent in its intended deployment setting. In a paired calibration, LLM judgments agreed with clinicians in 23--26 of 30 assessments across scored dimensions (quadratic-weighted $\kappa$ = 0.69--0.76; ICC = 0.69--0.77) and in all 30 hallucination checks (Appendix~\ref{subsec:llm_clinician_agreement}).

We additionally developed a web-based platform for user recruitment, user studies, and feedback collection, engaging more than 100 users. User feedback has been largely positive on the system's usefulness for case-report preparation. Detailed data-collection procedures, study protocols, blinded A/B results, and platform-based user feedback are provided in Appendix~\ref{subsec:human_evaluation_protocol}.

\section{Conclusion}
\label{sec:conclusion}
\vspace{-0.25\baselineskip}

\looseness=-1

We presented MedCaseAgent, a planner–writer–editor agent that turns fragmented clinical notes and medical images into publication-style rare-disease case reports, together with MedCase-150K and MedCase-Bench, a 584-case temporal holdout of reports published after the construction cutoff. Tool-grounded planning yields large gains in verifiable grounding: MedCaseAgent-Max retrieves the exact source article for 84.6\% of cases against 46.9\% for the strongest direct baseline, recognizes case novelty in 88.4\% of cases, and attains the highest must-cite citation recall. It is the best-performing system on every disease subset, from COVID-19 to ultra-rare diseases, and scores above the human-publication reference under a four-judge ensemble spanning four model families. In blinded review, experienced clinicians more often preferred MedCaseAgent reports over the published originals, and most did not consistently identify which report was generated. The remaining frontier is diseases with no precedent in the retrieval sources: there, MedCaseAgent can organize the available evidence and its uncertainty, but the discovery is the clinician's.

\label{lastmainpage}
\subsection*{AI Use Statement}

For manuscript preparation, generative AI tools were used for language polishing and improving readability. The authors reviewed the edits and take responsibility for the final manuscript. The use of LLMs for clinical-atom construction, planner-training trajectory generation, report generation, and automated evaluation is part of the research methodology described in Sections~\ref{sec:methodology} and~\ref{sec:experiments} and the appendices.

\subsection*{Ethics Statement}

The hospital-based case study part was conducted under existing institutional review board (IRB) approval, and all clinical materials were de-identified before analysis (Appendix~\ref{subsec:hospital_evaluation_protocol}). MedCaseAgent is intended to assist clinicians in preparing case reports. Generated reports require clinical review of patient facts, image interpretations, and citation support before dissemination. Broader impacts and responsible-use considerations are discussed in Appendix~\ref{subsec:broader_impacts}.

\subsection*{Reproducibility Statement}

We describe the method and evaluation setup in Sections~\ref{sec:methodology} and~\ref{sec:experiments}. Appendices~\ref{subsec:medcaseagent_workflow_details} and~\ref{subsec:medcase150k_data_construction_views} provide workflow and prompt displays, data-construction procedures, and validation details. Appendices~\ref{subsec:grpo_reward_sensitivity} and~\ref{subsec:compute_resources} report reward-weight ablations, training hyperparameters, and computational resources. Appendices~\ref{subsec:human_evaluation_protocol} and~\ref{subsec:llm_judged_ab_appendix} describe the human-evaluation and LLM-judge protocols. We plan to release the MedCase-150K and MedCase-Bench datasets, together with the source code, data-construction scripts, evaluation prompts, metadata, and data splits, as described in Appendix~\ref{subsec:release_llm_disclosure}. The hospital cases used for expert evaluation are non-public.

\subsection*{Acknowledgment}
MedCaseAgent was inspired by the authors' own experience in writing clinical case reports~\citep{zhu2025case,chen2024case}. PediaMed AI, a US-based startup, is supported in part by API credits from Anthropic’s AI for Science Rare Disease Program. Model access was restricted to US-based PediaMed AI employees, and all intellectual property arising from this work remains the sole property of PediaMed AI. Anthropic provided compute support only and was not involved in study design, data construction, evaluation, analysis, or content generation. The next version of MedCaseAgent will be powered by Claude Sonnet 5.5 and released alongside a platform.

\clearpage
\bibliography{main}
\bibliographystyle{bibstyle}


\clearpage
\appendix
\section*{Appendix}
\phantomsection\label{sec:appendix}
\crefalias{section}{appendix}
\renewcommand\thefigure{\Alph{section}\arabic{figure}}
\renewcommand\thetable{\Alph{section}\arabic{table}}

\startcontents[sections]
\printcontents[sections]{l}{1}{\setcounter{tocdepth}{2}}

\clearpage

\section{Extended Related Work: Relevant Benchmarks}
\label{sec:extended_related_work}
\setcounter{figure}{0}
\setcounter{table}{0}

Recent benchmarks for clinical agents can be grouped by what they evaluate. One group studies clinical reasoning and decision making, including HealthBench~\citep{arora2025healthbench}, MedAgentBench~\citep{jiang2025medagentbench}, MedConsultBench~\citep{qiao2026medconsultbench}, ClinDEF~\citep{tang2025clindef}, VivaBench~\citep{chiu2025simulating}, RareBench~\citep{chen2024rarebench}, and MedAgentBoard~\citep{zhu2025medagentboard}. A second group focuses on multimodal, imaging-centered, or sequential clinical decision-making evaluation, such as ReX-MLE~\citep{kenia2025rex}, MedicalNarratives~\citep{ikezogwo2025medicalnarratives}, AgentClinic~\citep{schmidgall2024agentclinic}, and MTBBench~\citep{vasilev2025mtbbench}. A third group targets clinical workflow, dialogue, and system-level behavior, including MedBench~\citep{cai2024medbench}, 3mdbench~\citep{sviridov20253mdbench}, Care-bench~\citep{wang2026care}, and MMRareBench~\citep{ning2026mmrarebench}. Other benchmarks examine safety, privacy, and structured extraction, such as MedPriv-Bench~\citep{guan2026medpriv}, MedRiskEval~\citep{corbeil2026medriskeval}, and ClinBench~\citep{villanueva2026clinbench}. PMC-Patients~\citep{zhao2023large} provides patient summaries for patient and literature retrieval, while RareArena~\citep{chen2026rarearena} benchmarks rare-disease diagnosis. Despite this growing benchmark ecosystem, most existing settings still focus on question answering, information extraction, summarization, diagnosis, report evaluation, or simulated interaction. Clinical case report generation brings these abilities together in a single task. A strong system must interpret fragmented multimodal evidence, reconstruct the patient course, link findings to prior literature, and write a rigorous scientific narrative.

\FloatBarrier
\clearpage

\section{Prompt and Workflow Displays}
\label{subsec:medcaseagent_workflow_details}
\setcounter{figure}{0}
\setcounter{table}{0}

\begin{figure}[p]
    \centering
    \includegraphics[width=\linewidth,height=0.82\textheight,keepaspectratio]{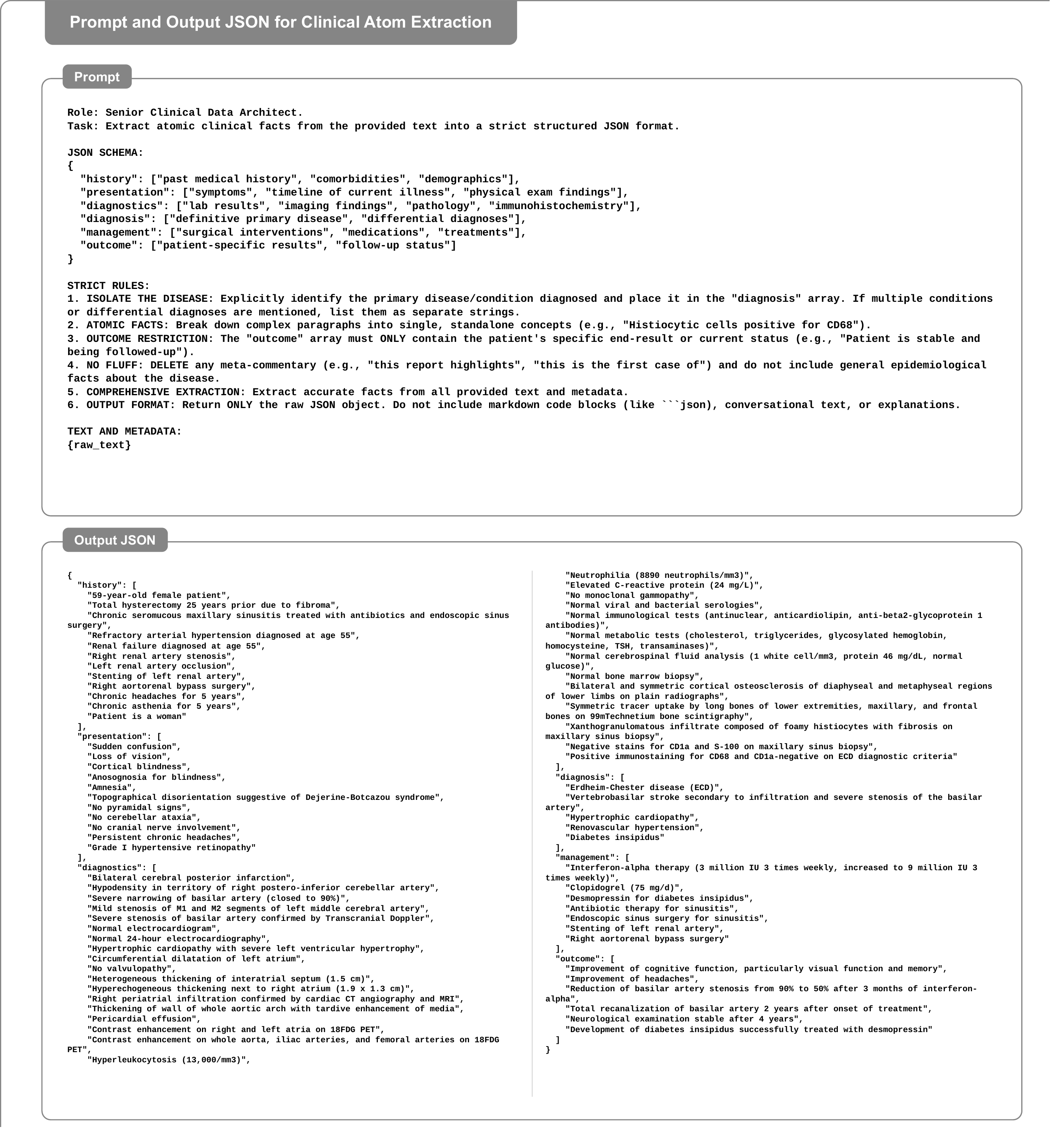}
    \caption{Clinical atom compression example. A polished source case report is compressed into fragmented, patient-specific atoms that resemble real-world shorthand notes. The prompt-output view illustrates how background prose and final diagnostic exposition are removed while preserving clinical history, findings, tests, interventions, outcomes, and image-relevant details.}
    \label{fig:appendix_prompt_output_atoms}
\end{figure}

\begin{figure}[htbp]
    \centering
    \includegraphics[width=\linewidth]{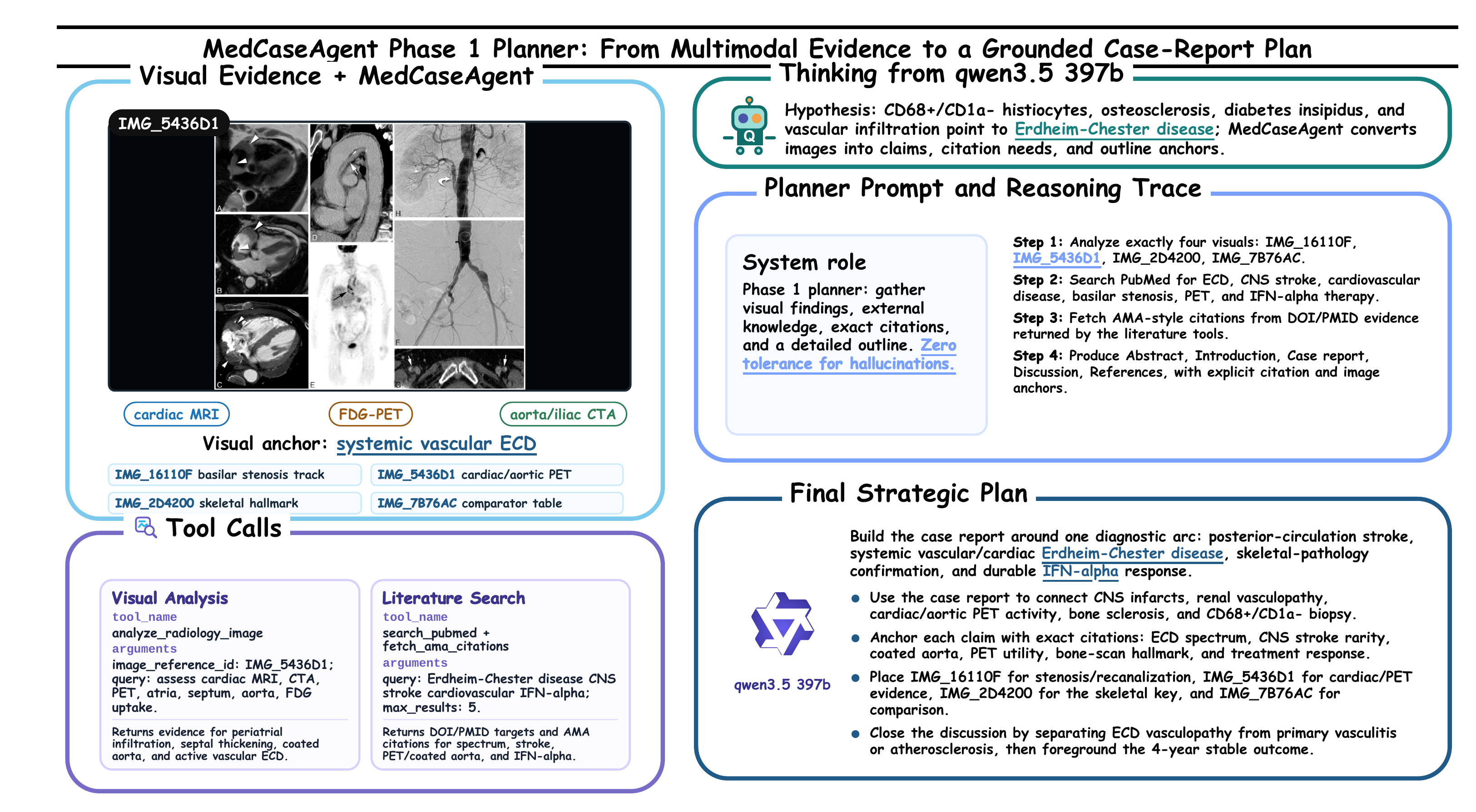}
    \caption{Detailed view of the MedCaseAgent planning phase. The planner organizes fragmented multimodal case evidence, retrieves relevant medical context, and produces a structured plan that guides downstream report writing and editing.}
    \label{fig:appendix_medcase_plan_phase}
\end{figure}

\begin{figure}[htbp]
    \centering
    \includegraphics[width=\linewidth]{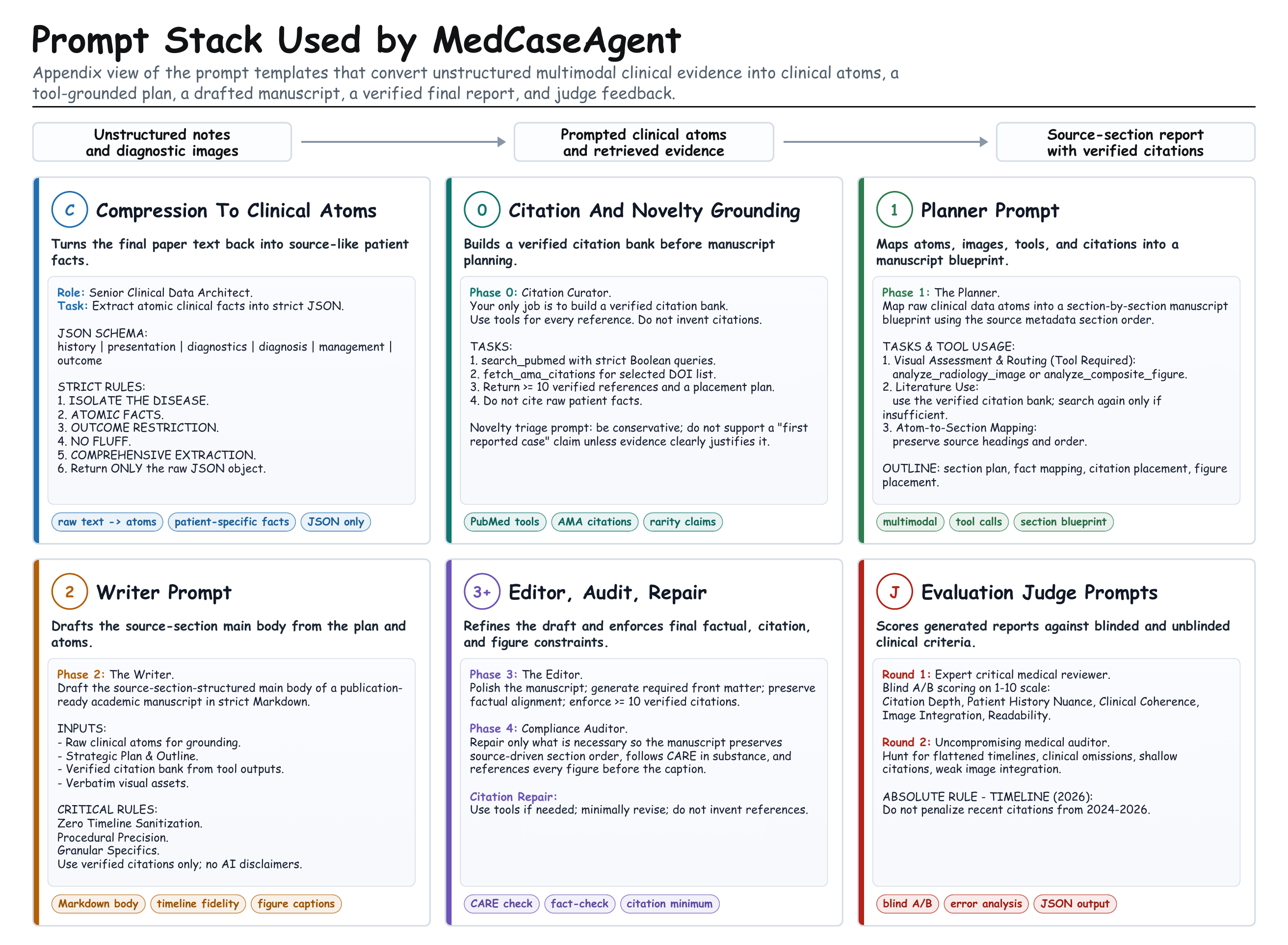}
    \caption{Prompt stack used by MedCaseAgent across the agentic report-generation workflow. The stack separates role instructions, case-specific evidence, retrieved context, output constraints, and audit requirements so that each phase receives grounded inputs for clinical case report synthesis.}
    \label{fig:appendix_medcaseagent_prompt_stack}
\end{figure}

This appendix gathers the prompt-facing displays for the agentic report-generation pipeline: clinical-atom compression, planner organization, and the prompt stack used across the workflow. Figure~\ref{fig:appendix_prompt_output_atoms} shows how polished source prose is reduced to patient-specific atoms before training and evaluation. Figure~\ref{fig:appendix_medcase_plan_phase} expands the planner stage that connects atoms, images, retrieved context, citations, and figure placement. Figure~\ref{fig:appendix_medcaseagent_prompt_stack} shows how instructions and evidence are separated across the agent workflow so that downstream drafting remains grounded in case-specific inputs.

\FloatBarrier
\clearpage

\section{Data Construction and Validation}
\label{subsec:medcase150k_data_construction_views}
\setcounter{figure}{0}
\setcounter{table}{0}

\subsection{Human Verification of Clinical Atoms and Rarity Labels}
\label{subsec:atom_human_verification}

During MedCase-Bench construction, 23 PhD researchers and domain experts with medical or biomedical backgrounds manually reviewed all 632 candidate cases. Each case was independently reviewed by at least two experts, with pre-adjudication judgments retained. The review proceeded in two stages: clinical-atom fidelity, followed by rarity/novelty-label validity for the cases retained after the first stage. Reviewers excluded cases that failed verification; they did not edit the compressed atoms or relabel the rarity categories.

For atom fidelity, reviewers checked each patient-specific fact against the source report for factual alterations, unsupported additions, temporal errors, and potentially misleading ambiguities. For rarity/novelty labels, they checked the source evidence against the predefined \texttt{rare\_disease}, \texttt{unprecedented\_presentation}, \texttt{atypical\_presentation}, and \texttt{standard\_case} categories.

\begin{table}[htbp]
    \centering
    \caption{Sequential human verification of clinical atoms and rarity/novelty labels during MedCase-Bench construction. Cases failing either stage were excluded rather than edited or relabeled; exclusion and retention percentages use the number reviewed at that stage. Each case was reviewed by at least two reviewers. Agreement is measured before adjudication.}
    \label{tab:clinical_atom_verification}
    \footnotesize
    \setlength{\tabcolsep}{3pt}
    \renewcommand\arraystretch{1.15}
    \rowcolors{2}{gray!10}{white}
    \begin{tabular}{@{}>{\raggedright\arraybackslash}p{0.25\linewidth}>{\centering\arraybackslash}p{0.12\linewidth}>{\centering\arraybackslash}p{0.18\linewidth}>{\centering\arraybackslash}p{0.18\linewidth}>{\centering\arraybackslash}p{0.19\linewidth}@{}}
        \toprule
        \rowcolor{gray!25}
        \textbf{Review stage} & \textbf{Reviewed ($N$)} & \textbf{Excluded $n$ (\%)} & \textbf{Retained $n$ (\%)} & \textbf{Pre-adjudication agreement} \\
        \midrule
        Clinical-atom fidelity & 632 & 30 (4.75\%) & 602 (95.25\%) & 94\% \\
        Rarity/novelty label & 602 & 18 (2.99\%) & 584 (97.01\%) & 87\% \\
        \bottomrule
    \end{tabular}
\end{table}

Table~\ref{tab:clinical_atom_verification} summarizes the sequential screening. Atom-fidelity review excluded 30 of 632 candidates, retaining 602 (95.25\%); rarity-label review excluded 18 of these 602 cases, retaining 584 (97.01\%). The final MedCase-Bench therefore contains 584 cases, with an overall retention rate of 92.41\% (584/632). Pre-adjudication agreement was 94\% for atom fidelity and 87\% for rarity labels. This procedure verifies fidelity to the published source; comparison with original hospital records is described separately in Appendix~\ref{subsec:hospital_evaluation_protocol}.

The disease-specific subsets in Table~\ref{tab:disease_category_scores} are sampled separately from MedCase-150K: each category uses 100 cases, except Erdheim--Chester disease (ECD), which uses all 53 available cases in that corpus.

\FloatBarrier

\subsection{Corpus Statistics and Data Construction Views}

\begin{figure}[htbp]
    \centering
    \includegraphics[width=\linewidth]{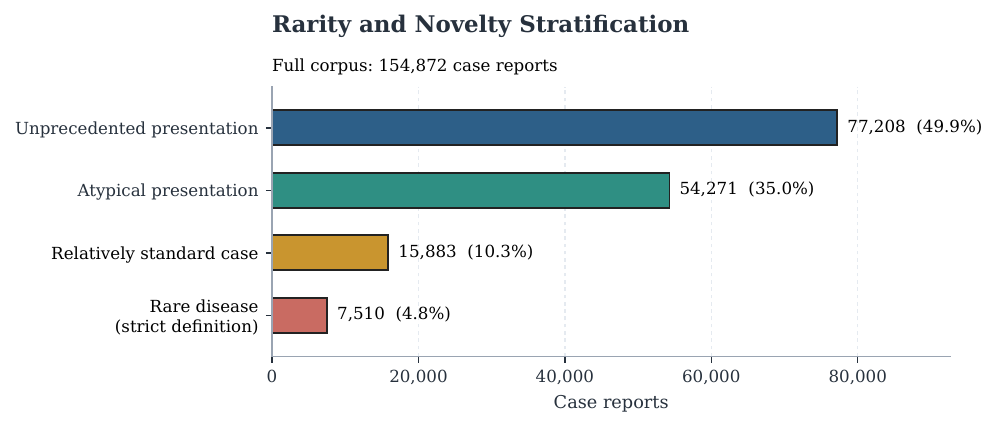}
    \caption{Rarity and novelty stratification for the curated case-report corpus. Bars summarize rarity labels assigned during data construction, separating rare diseases in the strict epidemiological sense, unprecedented presentations, atypical presentations, and relatively standard cases.}
    \label{fig:appendix_data_rarity_license}
\end{figure}

\begin{figure}[htbp]
    \centering
    \includegraphics[width=\linewidth]{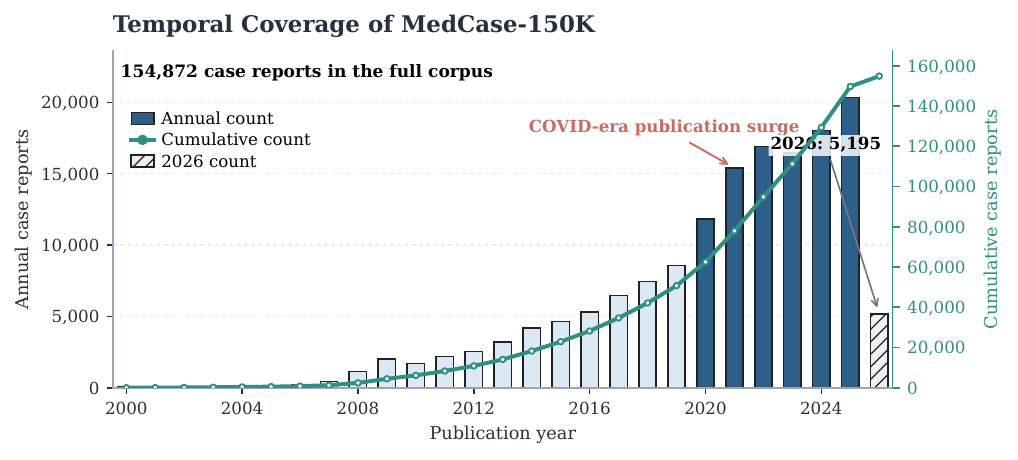}
    \caption{Temporal coverage of the MedCase-150K case-report corpus. Bars show annual case-report counts and the line shows cumulative coverage.}
    \label{fig:appendix_data_temporal}
\end{figure}

\begin{figure}[htbp]
    \centering
    \includegraphics[width=\linewidth]{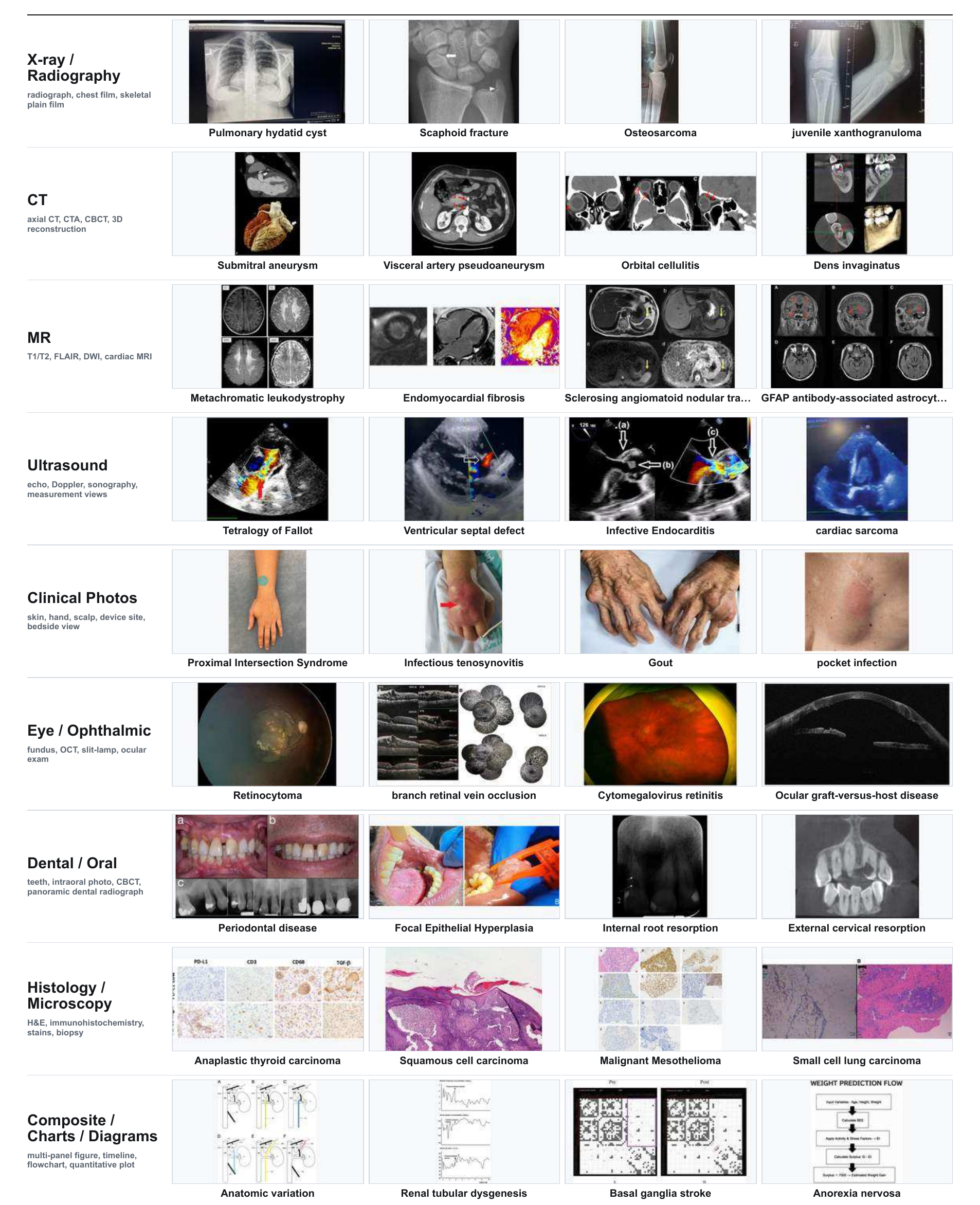}
    \caption{Representative visual complexity in MedCase-Bench. Rows group examples by imaging or figure modality, illustrating the breadth of radiology, clinical photography, ophthalmology, dentistry, pathology, and structured graphical evidence that the benchmark requires multimodal agents to interpret.}
    \label{fig:appendix_medcasebench_image_atlas}
\end{figure}

\begin{figure}[htbp]
    \centering
    \includegraphics[width=\linewidth]{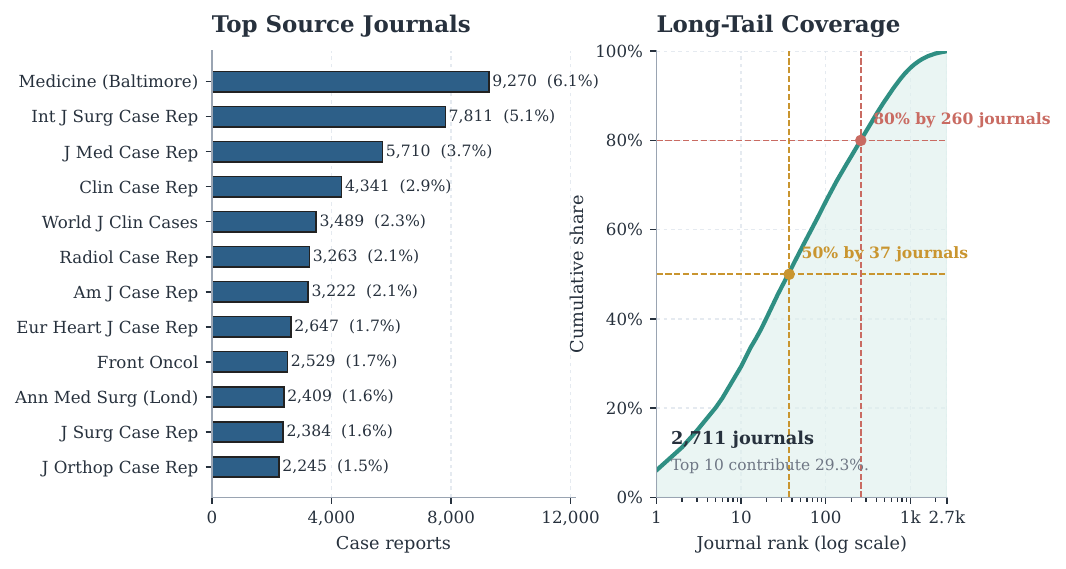}
    \caption{Source breadth across journals. The left panel reports the leading source venues, while the right panel shows that coverage remains broadly distributed across the long tail of journals.}
    \label{fig:appendix_data_sources}
\end{figure}

This section visualizes the corpus-level quality checks behind MedCase-150K and MedCase-Bench. Figure~\ref{fig:appendix_data_rarity_license} summarizes the rarity labels assigned during data construction. Figure~\ref{fig:appendix_data_temporal} shows the temporal coverage used for the post-January-2026 evaluation split. Figure~\ref{fig:appendix_medcasebench_image_atlas} samples the visual modalities that make the benchmark multimodal. Figure~\ref{fig:appendix_data_sources} checks whether the corpus is concentrated in a few journals or distributed across many source venues.

\paragraph{Rarity and novelty classification} Rare disease status is not always a stable binary property: the prevalence, clinical significance, and publication novelty of a condition can change across time, geography, and medical context. For publications identified as case reports, we assign one of four rarity labels. \texttt{rare\_disease} means that the underlying disease or condition is epidemiologically rare. \texttt{unprecedented\_presentation} means that the report describes a first-of-its-kind or completely novel manifestation. \texttt{atypical\_presentation} means that a known disease appears in an uncommon form, location, demographic group, or clinical context. \texttt{standard\_case} means that the report is a useful teaching case but does not contain a major rare-disease or novel-presentation element. For stratified corpus views, \texttt{rare\_disease}, \texttt{unprecedented\_presentation}, and \texttt{atypical\_presentation} are grouped as rare or novelty-bearing case reports, while \texttt{standard\_case} is retained as the non-rare/non-novel comparison group.

\FloatBarrier
\clearpage

\section{GRPO Reward-Weight Sensitivity}
\label{subsec:grpo_reward_sensitivity}
\setcounter{figure}{0}
\setcounter{table}{0}

\paragraph{Reward design and weight selection}
Each of the six reward components is bounded to $[0,1]$ before aggregation, and the default training configuration uses deterministic rewards without invoking a live LLM judge. The default weights form a two-tier heuristic based on the task objectives. Exact DOI retrieval, evidence support, clinical-atom/section coverage, and the overall quality proxy represent evidence retrieval, evidential validity, coverage of patient facts, and plan quality, respectively, and each receives 0.20. Tool-schema compliance and structural/image-placement constraints are necessary for valid outputs but tend to saturate early; each receives 0.10 so that formatting rewards do not outweigh clinical content and evidence.

\paragraph{Experimental setup}
We evaluate sensitivity by repeating GRPO from the same SFT checkpoint under three alternative normalized weight profiles, holding the training data, optimization budget, sampling configuration, and evaluation procedure fixed. Uniform assigns equal weight to all six components; Retrieval-heavy increases exact DOI retrieval and evidence support; and Quality/image-heavy increases the quality proxy and image placement. A separate No-DOI ablation removes the source-DOI reward and renormalizes the remaining weights. In the component order $[\mathrm{schema},\mathrm{doi},\mathrm{evidence},\mathrm{coverage},\mathrm{image},\mathrm{quality}]$, the five evaluated profiles are
{
\begin{equation*}
\begin{aligned}
\mathbf{w}_{\text{Default}} &= [0.10,\,0.20,\,0.20,\,0.20,\,0.10,\,0.20],\\
\mathbf{w}_{\text{Uniform}} &= [\tfrac{1}{6},\,\tfrac{1}{6},\,\tfrac{1}{6},\,\tfrac{1}{6},\,\tfrac{1}{6},\,\tfrac{1}{6}],\\
\mathbf{w}_{\text{Retrieval-heavy}} &= [0.10,\,0.30,\,0.30,\,0.10,\,0.10,\,0.10],\\
\mathbf{w}_{\text{Quality/image-heavy}} &= [0.10,\,0.10,\,0.10,\,0.10,\,0.30,\,0.30],\\
\mathbf{w}_{\text{No-DOI}} &= [0.125,\,0,\,0.25,\,0.25,\,0.125,\,0.25].
\end{aligned}
\end{equation*}
}

\begin{table}[htbp]
    \centering
    \caption{GRPO reward-weight sensitivity from the same SFT checkpoint, with training data, optimization budget, sampling configuration, and evaluation procedure held fixed. Bold indicates the best result in each column.}
    \label{tab:grpo_reward_sensitivity}
    \small
    \setlength{\tabcolsep}{4pt}
    \renewcommand\arraystretch{1.08}
    \rowcolors{2}{gray!10}{white}
    \begin{adjustbox}{max width=\linewidth}
    \begin{tabular}{lccccc}
        \toprule
        \rowcolor{gray!25}
        \textbf{Weight setting} & \tableheadercell{\textbf{Atom}\\\textbf{coverage}} & \tableheadercell{\textbf{Image}\\\textbf{placement}} & \tableheadercell{\textbf{Novelty}\\\textbf{recognition}} & \tableheadercell{\textbf{Report}\\\textbf{quality}} & \textbf{Average} \\
        \midrule
        Default & \textbf{86.30} & 74.12 & 59.08 & \textbf{67.23} & \textbf{71.68} \\
        Uniform & 85.08 & 73.95 & 58.21 & 66.47 & 70.93 \\
        Retrieval-heavy & 84.79 & 72.49 & \textbf{60.02} & 66.48 & 70.95 \\
        Quality/image-heavy & 83.25 & \textbf{74.63} & 57.85 & 66.86 & 70.65 \\
        \midrule
        No DOI reward ($w_{\mathrm{doi}}=0$) & 84.26 & 73.91 & 55.78 & 65.37 & 69.83 \\
        \bottomrule
    \end{tabular}
    \end{adjustbox}
\end{table}

\paragraph{Results}
Table~\ref{tab:grpo_reward_sensitivity} reports the average of atom coverage, image placement, novelty recognition, and report quality. The default weights achieve the highest average (71.68), atom coverage, and report quality. Retrieval-heavy leads novelty recognition, while Quality/image-heavy leads image placement. Across the four profiles retaining the source-DOI reward, the averages range from 70.65 to 71.68, a spread of 1.03 points.

\paragraph{Removing the source-DOI reward}
To assess the contribution of $R_{\mathrm{doi}}$, the No-DOI profile sets $w_{\mathrm{doi}}=0$ and renormalizes the default weights of the remaining components. Its four-metric average is 69.83, compared with 71.68 for Default. Novelty recognition decreases by 3.30 percentage points (59.08\% to 55.78\%), and report quality decreases by 1.86 points (67.23 to 65.37). These results support a contribution of the source-DOI reward to novelty recognition and report quality under the evaluated training configuration.

\clearpage

\section{Training and Compute Resources}
\label{subsec:compute_resources}
\setcounter{figure}{0}
\setcounter{table}{0}

The tuned MedCaseAgent planner reported in the experiments is initialized from Qwen3.5-9B and trained with a two-stage SFT-to-GRPO recipe using the SWIFT training stack. The planner training set contains 3,000 phase-1 planning trajectories derived from the MedCase-150K training split. These trajectories are generated by a Qwen3.5-397B-A17B teacher planner: for each case, the teacher reads the compressed clinical atoms and mapped images, issues image-analysis and literature-search actions, records public tool observations, builds an evidence table with DOI-linked citations, and writes the final section-level case-report plan. The data-preparation step converts each teacher trace into SWIFT agent messages and removes private thinking fields. For the GRPO and evaluation records, source DOI identifiers are kept only as scorer-side labels; if search hits the source article for a case, its title, abstract, and article text are masked from policy-visible observations, while non-source PubMed records remain available as retrieved support.

We train LoRA adapters rather than full model weights, targeting the attention and MLP projection modules ($q$, $k$, $v$, $o$, gate, up, and down projections) with rank 16, alpha 32, and dropout 0.05. Training uses 4-bit bitsandbytes quantization, bfloat16 compute, and gradient checkpointing.

\begin{center}
\small
\begin{tabular}{p{0.30\linewidth}p{0.62\linewidth}}
\toprule
Item & Setting \\
\midrule
Base model & Qwen3.5-9B \\
Trajectory teacher & Qwen3.5-397B-A17B \\
Training data & 3,000 teacher-generated planning trajectories from the training split \\
Hardware & 8 NVIDIA H100 GPUs, 80GB memory each \\
Wall-clock time & Approximately one day for the final planner adaptation run \\
Estimated compute (planner adaptation only) & Approximately 192 H100 GPU-hours \\
SFT objective & Teacher-forced planner trajectory learning \\
SFT effective batch & 128 examples (8 GPUs $\times$ batch 1 $\times$ accumulation 16) \\
SFT epochs & 2 \\
SFT learning rate & $2.0\times10^{-5}$ with cosine decay and 0.03 warmup ratio \\
GRPO rollout group & 4 candidate planner trajectories per prompt \\
GRPO effective prompt batch & 64 prompts per update (8 GPUs $\times$ batch 1 $\times$ accumulation 8) \\
GRPO maximum steps & 500 optimizer updates \\
GRPO sampling & Temperature 0.8, top-$p$ 0.95, KL coefficient $\beta=0.02$ \\
GRPO learning rate & $5.0\times10^{-6}$ with cosine decay and 0.03 warmup ratio \\
GRPO rewards & Weighted sum of schema, DOI retrieval, evidence support, atom coverage, image placement, and quality proxy rewards \\
\bottomrule
\end{tabular}
\end{center}

The compute estimate covers the reported SFT+GRPO LoRA adaptation run and does not include teacher-model inference, preliminary prompt or tool development, or inference-only evaluation.

The GRPO stage follows the configured 500-step cap and scores each sampled trajectory with
\begin{equation}
R(\tau)=0.10R_{\mathrm{schema}}+0.20R_{\mathrm{doi}}+0.20R_{\mathrm{evidence}}+0.20R_{\mathrm{coverage}}+0.10R_{\mathrm{image}}+0.20R_{\mathrm{quality}},
\end{equation}
where all component scores are clipped to $[0,1]$ before aggregation. The default training path uses deterministic rewards and does not call a live judge. The reward components and weights used by the planner adaptation are:

\begin{center}
\small
\begin{tabular}{p{0.24\linewidth}p{0.10\linewidth}p{0.54\linewidth}}
\toprule
Component & Weight & Reward signal \\
\midrule
Tool schema & 0.10 & Valid tool names, required arguments, valid image identifiers for image-analysis calls, DOI-bearing citation-fetch calls, and non-duplicate tool calls. \\
Exact DOI retrieval & 0.20 & Recovery of scorer-only source DOI labels when available, with source title, abstract, and article text withheld from policy-visible retrieval context. \\
Evidence support & 0.20 & DOI format validity, citation-anchor validity, and overlap between generated citations and non-source retrieved support DOIs. \\
Plan coverage & 0.20 & Coverage of clinical atom groups and required report sections, reasonable plan length, and plan-like formatting. \\
Structure and images & 0.10 & Parse success, required-section coverage, placement of provided image identifiers, and citation-anchor validity. \\
Quality proxy & 0.20 & Deterministic blend of structure, evidence, coverage, explicit gap awareness, and avoiding unsupported citation-from-memory patterns; a precomputed judge score can replace this proxy when supplied. \\
\bottomrule
\end{tabular}
\end{center}

\clearpage

\section{Clinical Evaluation and User Study Details}
\label{subsec:human_evaluation_protocol}
\setcounter{figure}{0}
\setcounter{table}{0}

Section~\ref{subsec:clinical_evaluation_user_study} summarizes the human evaluation. This appendix describes hospital-case collection and expert assessment, differences between raw clinical records and report-derived atoms, paired LLM--clinician agreement, the blinded A/B study, and the web-based user studies.

\subsection{Hospital-Based Expert Evaluation}
\label{subsec:hospital_evaluation_protocol}

\paragraph{Case collection}
We retrospectively assembled 30 original, non-public, de-identified multimodal cases from 10 hospitals through an ongoing case-report effort coordinated by the authors' institution under existing IRB approval. Each case involved a rare or clinically complex presentation and included longitudinal clinical records, multimodal evidence, and a corresponding report prepared and reviewed by the clinical team. These reports included both published articles and complete unpublished manuscripts prepared for submission. Cases were selected for sufficiently complete records and a matched clinician-written report; consolidating records across hospitals and care episodes required clinical coordination and manual review. All materials were de-identified before analysis.

\paragraph{Inputs and source verification}
For each patient, MedCaseAgent generated one report directly from the raw clinical records ($E_{\mathrm{raw}}$) and another from report-derived clinical atoms ($E_{\mathrm{atom}}$). The atoms were obtained by compressing the clinician-written manuscript, rather than by directly extracting facts from the raw clinical files; images were supplied separately without captions or explicit atom-level links. Experts compared both generated reports with the clinician-written report and used the original records to verify patient facts. Raw records contained redundancy, irregular timelines, uncertainty, and cross-source conflicts, requiring filtering, deduplication, source reconciliation, and timeline reconstruction. Report-derived atoms retained the core history, examinations, treatment, and follow-up in a more concise form, but compressed some source context.

\paragraph{Expert assessment}
Fourteen clinical experts assessed the 30 cases, with every case independently reviewed by three experts. Experts assessed overall report quality, patient-timeline reconstruction, and omission control on a 1--10 scale, with higher scores indicating better performance, while checking the reports against the source records. Table~\ref{tab:hospital_expert_evaluation} reports the mean scores for raw-record generation, atom-based generation, and clinician-written reports.

\paragraph{Raw-record versus atom-based generation}
In the order of overall quality, patient timeline, and omission control, mean scores were 7.4/7.5/7.8 for raw-record generation, 7.6/7.6/7.6 for atom-based generation, and 7.5/7.8/7.9 for clinician-written reports. Atom-based generation received slightly higher quality and timeline scores than raw-record generation, while raw records provided better omission control. Both input forms achieve quality on par with clinician-written reports, supporting atoms as a controlled benchmark input whose results carry over to real hospital records.

\paragraph{Same-patient input comparison}
Table~\ref{tab:raw_atom_case_comparison} illustrates the input differences using a de-identified pediatric cerebral visual impairment case. Its original source package comprised two DOCX files, three PDFs totaling 76 pages, and seven standalone PNG images. The corresponding clinician-written manuscript contained four embedded figures. The comparison shows how report-derived atoms retain central patient facts while compressing temporal context, source provenance, conflicting records, and acquisition differences. The model must still associate images, organize evidence, retrieve literature, and construct a report, while raw inputs additionally require document filtering and cross-source reconciliation.

\begin{table}[htbp]
    \centering
    \caption{Same-patient example of atomization for visual dysfunction following encephalitis. The atoms were derived from a clinician-written manuscript, not directly from the original clinical files. The abbreviated comparisons show source context compressed by the corresponding report-derived atoms.}
    \label{tab:raw_atom_case_comparison}
    \footnotesize
    \setlength{\tabcolsep}{4pt}
    \renewcommand\arraystretch{1.18}
    \rowcolors{2}{gray!10}{white}
    \begin{tabular}{@{}>{\raggedright\arraybackslash}p{0.34\linewidth}>{\raggedright\arraybackslash}p{0.33\linewidth}>{\raggedright\arraybackslash}p{0.27\linewidth}@{}}
        \toprule
        \rowcolor{gray!25}
        \textbf{Raw documentation} & \textbf{Corresponding report-derived atom} & \textbf{Information compressed} \\
        \midrule
        Dated entries separately record fever, marrow findings, impaired consciousness, visual loss, and later motor milestones. &
        Visual dysfunction persisting for over four months after EBV-associated encephalitis. &
        The central presentation is retained, while earlier events, care transitions, and temporal context are compressed. \\
        Visual findings occur in narrative text and photographed reports. Several BAEP, P-VEP, and OCT screenshots contain demographic details inconsistent with the index patient. &
        P-VEP showed bilateral NPN waveforms with prolonged latencies; fundoscopy showed no obvious structural abnormality. &
        Summarized findings are retained, while provenance and possible wrong-patient contamination may be lost. \\
        A summary specifies occipital rTMS at 90--100\% motor threshold and 5--6 Hz. The contemporaneous plan records 20\% intensity, a 25\% threshold, 5 Hz, and 750 stimuli. &
        Occipital rTMS at 90--100\% resting motor threshold, 5--6 Hz, and 20 minutes per day. &
        A single-version atom may obscure conflicting treatment parameters across sources. \\
        Left/right optic-radiation streamline counts are 20/496, 314/21, and 483/10 across scans. The final diffusion scan changes from 3 T to 7 T. &
        Left optic-radiation fiber count increased, whereas the right decreased. &
        A direction-only atom may omit acquisition differences between scans. \\
        \bottomrule
    \end{tabular}
\end{table}

\FloatBarrier

\subsection{LLM--Clinician Agreement}
\label{subsec:llm_clinician_agreement}

\paragraph{Paired calibration}
We report agreement with clinical experts separately from the LLM-judged comparison of baseline systems. The alignment statistics below come from a calibration on 10 hospital cases: five clinical experts each reviewed six cases, giving 30 clinician--case assessments, with three independent reviews per case. Each assessment was paired with an LLM-judge assessment of the same output under a shared rubric. These 30 paired assessments are distinct from the aggregate results of the expanded 30-case, 14-expert hospital evaluation in Appendix~\ref{subsec:hospital_evaluation_protocol}.

\paragraph{Agreement criteria}
Factuality, clinical coherence, omission, and citation accuracy were scored on 10-point scales. A pair agreed when the absolute difference between the LLM and clinician scores was at most two points. Hallucination used a binary rule: any hallucinated content constituted a failure, and no hallucination constituted a pass. Table~\ref{tab:llm_clinician_agreement} reports agreement counts and the correlation summaries for the scored dimensions; the hallucination row reports exact agreement.

\paragraph{Chance-corrected agreement}
Table~\ref{tab:llm_clinician_agreement} adds quadratic-weighted Cohen's $\kappa$ and two-way, absolute-agreement, single-measure ICC(A,1), computed between the paired LLM and clinician score columns. These are descriptive point estimates over the 30 assessments; repeated cases and clinicians are not modeled, and no cluster-adjusted intervals are reported. The previously reported agreement counts and correlation summaries are retained. Both added coefficients are undefined for hallucination because all ratings occupy the same category.

\begin{table}[htbp]
    \centering
    \caption{LLM--clinician agreement in the initial paired calibration (30 clinician--case assessments). Agreement on scored dimensions allows an absolute difference of at most two points on a 10-point scale. Hallucination uses binary pass/fail ratings and exact agreement.}
    \label{tab:llm_clinician_agreement}
    \small
    \setlength{\tabcolsep}{4pt}
    \renewcommand\arraystretch{1.12}
    \rowcolors{2}{gray!10}{white}
    \begin{tabular}{>{\raggedright\arraybackslash}p{\dimexpr.28\linewidth-2\tabcolsep\relax}
        >{\centering\arraybackslash}p{\dimexpr.14\linewidth-2\tabcolsep\relax}
        >{\centering\arraybackslash}p{\dimexpr.30\linewidth-2\tabcolsep\relax}
        >{\centering\arraybackslash}p{\dimexpr.14\linewidth-2\tabcolsep\relax}
        >{\centering\arraybackslash}p{\dimexpr.14\linewidth-2\tabcolsep\relax}}
        \toprule
        \rowcolor{gray!25}
        \textbf{Evaluation dimension} & \textbf{Agreement} & \tableheadercell{\textbf{Reported correlation /}\\\textbf{exact agreement}} & \tableheadercell{\textbf{Weighted}\\$\boldsymbol{\kappa}$} & \textbf{ICC} \\
        \midrule
        Factuality & 23/30 & $0.76 \pm 0.06$ & 0.74 & 0.74 \\
        Clinical coherence & 24/30 & $0.78 \pm 0.05$ & 0.76 & 0.77 \\
        Omission & 26/30 & $0.71 \pm 0.06$ & 0.69 & 0.69 \\
        Hallucination & 30/30 & 1.00 & N/A & N/A \\
        Citation accuracy & 24/30 & $0.74 \pm 0.06$ & 0.73 & 0.74 \\
        \bottomrule
    \end{tabular}
    \par\smallskip
    {\footnotesize\raggedright ICC: two-way absolute agreement, single measure, ICC(A,1). N/A: undefined because both rating columns are constant.\par}
\end{table}

Agreement was observed in 23/30 pairs for factuality, 24/30 for clinical coherence, 26/30 for omission, and 24/30 for citation accuracy. For hallucination, both raters assigned a pass in all 30 pairs, giving exact agreement of 1.00. This alignment supports using the LLM judge ensemble for large-scale comparison across baselines, complemented by clinician evaluation.

\FloatBarrier

\subsection{Expert-Reviewer Blind A/B Test}
\label{subsec:clinician_blinded_validation}

Because claims about medical AI value require evidence that goes beyond technical performance alone (``Show us the evidence for the value of medical AI'', \citeyear{ref1}), we tested whether automatic gains translate into real manuscript-review value through a blinded A/B study. In this study, 32 expert clinicians (with at least 5 years of review experience in clinical case reports) each reviewed five paired cases, yielding 160 pairwise judgments. In each pair, the clinician saw the same de-identified case context and images, a MedCaseAgent-Max report and the corresponding published reference in randomized order, then judged authorship and selected the stronger case report. We summarize source-identification consistency at the reviewer level by counting clinicians who correctly identify the generated report in at most three of five pairs. Under this criterion, 26 of 32 clinicians (81.3\%) did not consistently identify report sources across their five assigned pairs; reviewers also more often selected MedCaseAgent as the stronger manuscript.

\paragraph{Study design}
The paired cases are drawn from the post-January-2026 evaluation split, and report source labels are hidden from reviewers. The interface asks two primary forced-choice questions---which report appears to have been generated by an LLM, and which report the clinician prefers overall---to separate authorship indistinguishability from practical clinical-writing preference. Reviewers may also enter optional 1--5 rubric scores and free-text comments on hallucination, image interpretation, and overall quality. We report only aggregate statistics and omit reviewer identifiers, raw comments, and patient-level content.

\paragraph{Endpoint definition}
The primary human endpoint is a descriptive measure of source-identification consistency among experienced clinicians across the five-case blinded round. Independent random guessing meets the same threshold with probability 81.25\%. Under this criterion, most experienced clinicians did not consistently distinguish generated reports from published references across their five assigned pairs. Manuscript preference is used as complementary evidence of practical utility, and optional rubric scores provide additional descriptive detail.

\begin{figure}[p]
    \centering
    \includegraphics[width=\linewidth,height=0.76\textheight,keepaspectratio]{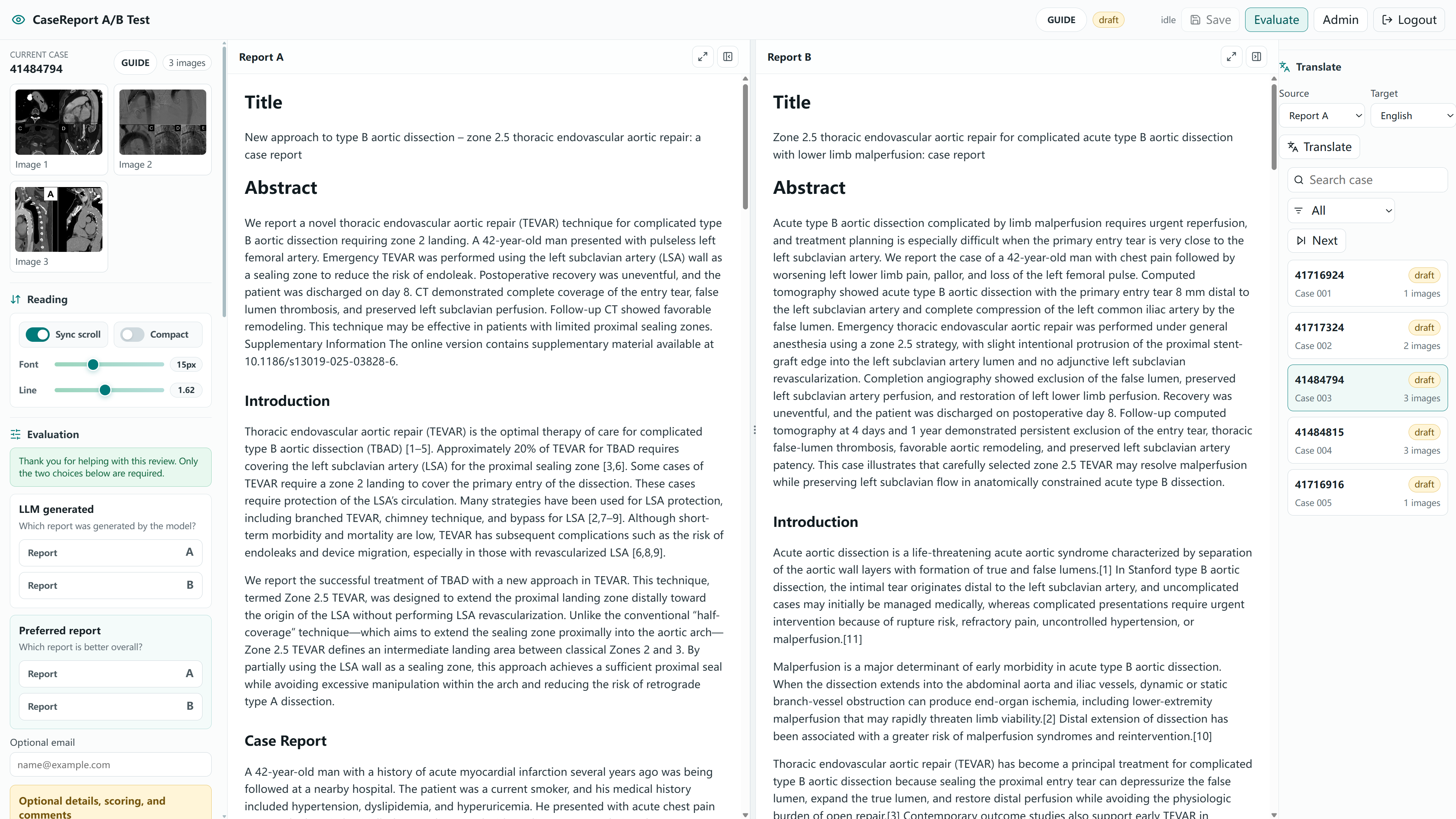}
    \caption{Expert web interface for the blinded A/B user study. Clinical reviewers inspect paired reports with shared medical images, guess which report was generated by an LLM, choose the stronger report overall, and may provide rubric-aligned scores and comments.}
    \label{fig:appendix_user_study_interface}
\end{figure}

\paragraph{Interface}
Figure~\ref{fig:appendix_user_study_interface} documents the reviewer-facing interface used to collect the blinded comparisons, showing how the shared case context, paired reports, and medical images are presented before clinicians choose the more likely generated report, select the preferred report, and optionally enter rubric-aligned scores.

\subsection{Web-Based User Studies and Feedback}
\label{subsec:platform_user_feedback}

We developed a web-based platform for user recruitment, user studies, and feedback collection, engaging more than 100 users. The platform supports study participation and collection of feedback on the system's usefulness for case-report preparation. For the expert A/B study, the reviewer-facing interface shown in Figure~\ref{fig:appendix_user_study_interface} records authorship judgments, report preferences, optional rubric scores, and free-text comments. User feedback has been largely positive on the system's usefulness for preparing case reports.

\FloatBarrier
\clearpage

\section{LLM-Judged Blind A/B Comparison Across Baselines}
\label{subsec:llm_judged_ab_appendix}
\setcounter{figure}{0}
\setcounter{table}{0}

\begin{table*}[!t]
    \centering
    \caption{LLM-judged blind A/B testing averaged across Gemini 3.1 Pro, GPT-5.4, Qwen3.5-397B-A17B, and Claude Opus 4.6 judges. All entries are reported as percentages.}
    \label{tab:blind_ab_test}
    \footnotesize
    \setlength{\tabcolsep}{4pt}
    \renewcommand\arraystretch{1.15}
    \rowcolors{2}{gray!10}{white}
      \begin{tabular}{lccc}
        \toprule
        \rowcolor{gray!25}
        \textbf{Method} & \tableheadercell{\textbf{Identified as}\\\textbf{Generated (\%)} $\downarrow$} & \tableheadercell{\textbf{Identified as}\\\textbf{Human (\%)} $\uparrow$} & \tableheadercell{\textbf{Preference vs.}\\\textbf{Ground Truth (\%)}} \\
        \midrule
        Qwen3.5-9B & 86.53 & 13.47 & 24.84 \\
        Gemma-4-31B & 82.31 & 17.69 & 30.58 \\
        Claude Opus 4.6 & 72.18 & 27.82 & 45.28 \\
        GPT-5.4 & 70.24 & 29.76 & 53.70 \\
        Gemini 3.1 Pro & 71.07 & 28.93 & 50.48 \\
        \midrule
        MedCaseAgent-Base & 79.67 & 20.33 & 36.29 \\
        MedCaseAgent-Tuned & 75.54 & 24.46 & 43.87 \\
        MedCaseAgent-Max & 63.91 & 36.09 & 64.26 \\
        \bottomrule
      \end{tabular}
\end{table*}

Table~\ref{tab:blind_ab_test} extends the MedCaseAgent-Max-vs.-reference comparison reported with human judges in Appendix~\ref{subsec:clinician_blinded_validation} to all baseline systems, using an ensemble of four LLM judges (Gemini 3.1 Pro, GPT-5.4, Qwen3.5-397B-A17B, Claude Opus 4.6) under the same blinded-A/B protocol with hidden model names, randomized A/B order, and evidence-grounded rationales \citep{liu2023g,zheng2023judging}. The ensemble enables coverage across all baselines at a scale that would be infeasible with clinicians. Paired LLM--clinician agreement is reported separately in Appendix~\ref{subsec:llm_clinician_agreement}. MedCaseAgent-Max is the least often identified as generated ($63.91\%$) and the most often preferred over the reference ($64.26\%$), with MedCaseAgent-Base and MedCaseAgent-Tuned showing stepwise gains over the non-agentic baselines. The clinician study complements this automatic table: most expert reviewers fail to recover MedCaseAgent's source labels across a full blinded review round while still finding the generated manuscripts useful, supporting the claim that the system is strong under automatic evaluation and persuasive in clinical review.

\clearpage
\subsection{Judge Robustness and Self-Preference}
\label{subsec:judge_robustness}

Three judges (GPT-5.4, Gemini 3.1 Pro, and Claude Opus 4.6) also serve as backbones in the configuration comparison. Table~\ref{tab:judge_robustness} reports general-quality scores under individual judges and the four-judge average.

\begin{table}[htbp]
    \centering
    \caption{General-quality scores under individual judges and the four-judge average.}
    \label{tab:judge_robustness}
    \small
    \setlength{\tabcolsep}{4pt}
    \renewcommand\arraystretch{1.18}

    \begin{tabular}{>{\raggedright\arraybackslash}p{\dimexpr.30\linewidth-2\tabcolsep\relax}*{5}{>{\centering\arraybackslash}p{\dimexpr.14\linewidth-2\tabcolsep\relax}}}
        \toprule
        \rowcolor{gray!25}
        \textbf{System} & \textbf{Qwen} & \textbf{Gemini} & \textbf{Claude} & \textbf{GPT} & \textbf{Average} \\
        \midrule
        Human Publication & 76.74 & 78.35 & 74.85 & 72.13 & 75.52 \\
        GPT-5.4 & 77.65 & 78.32 & 75.84 & 74.78 & 76.65 \\
        Gemini 3.1 Pro & 76.14 & 78.83 & 74.96 & 72.75 & 75.67 \\
        Claude Opus 4.6 & 75.60 & 77.02 & 75.10 & 70.84 & 74.64 \\
        MedCaseAgent-Base & 66.24 & 67.70 & 64.36 & 62.42 & 65.18 \\
        MedCaseAgent-Max & 84.10 & 85.48 & 82.34 & 80.76 & 83.17 \\
        \bottomrule
    \end{tabular}

\end{table}

\FloatBarrier
\clearpage

\section{Example Case Reports Generated by MedCaseAgent}
\label{subsec:generated_case_report_sheets}
\setcounter{figure}{0}
\setcounter{table}{0}

\begin{figure}[p]
    \centering
    \includegraphics[width=\linewidth,height=0.76\textheight,keepaspectratio]{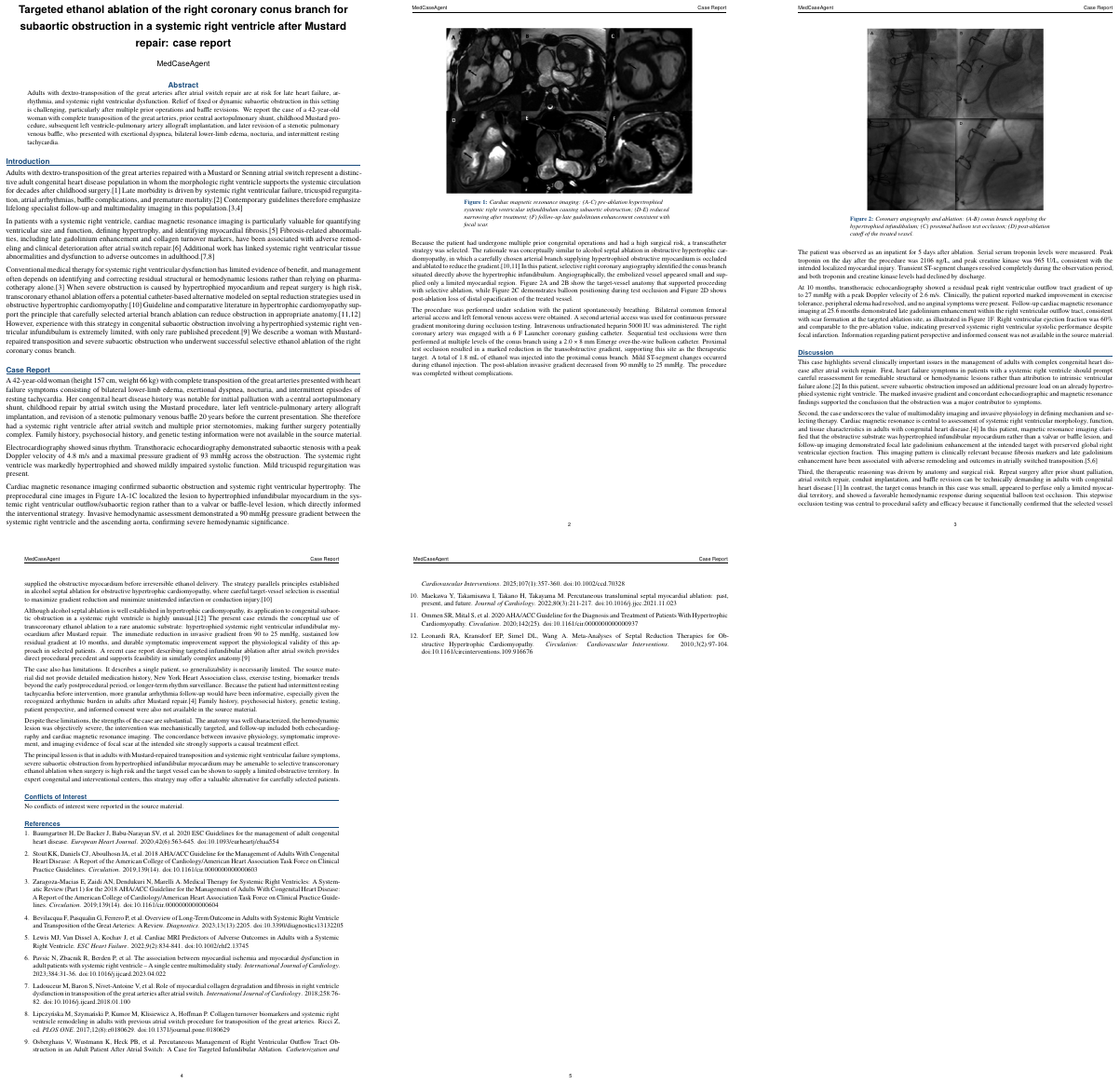}
    \caption{Example 1 of a case report generated by MedCaseAgent. The full-page sheet shows the generated manuscript structure, image placement, citation formatting, and clinical timeline organization.}
    \label{fig:appendix_generated_report_sheet_1}
\end{figure}

\begin{figure}[p]
    \centering
    \includegraphics[width=\linewidth,height=0.76\textheight,keepaspectratio]{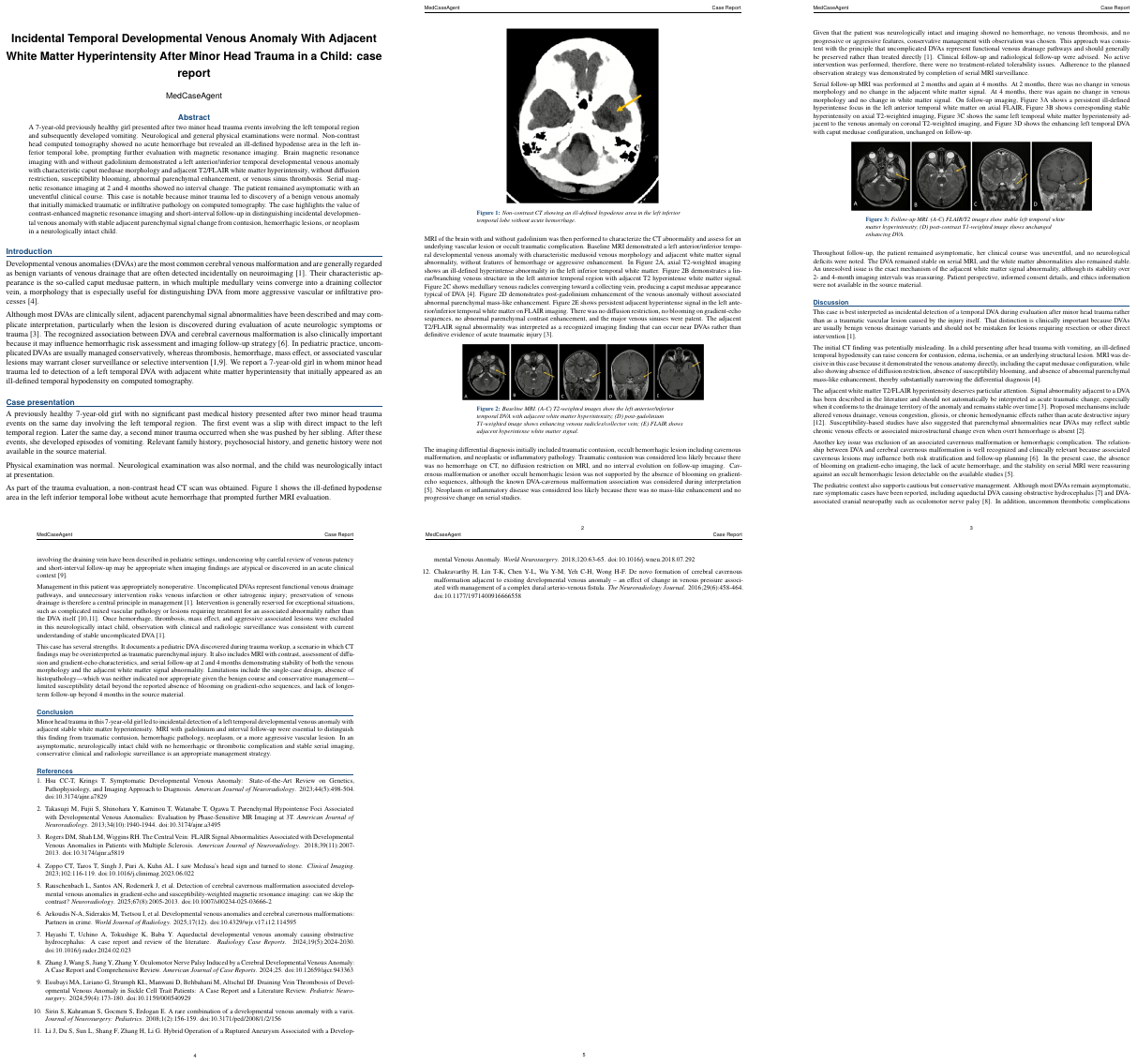}
    \caption{Example 2 of a case report generated by MedCaseAgent. The full-page sheet shows the generated manuscript structure, image placement, citation formatting, and clinical timeline organization.}
    \label{fig:appendix_generated_report_sheet_2}
\end{figure}

\begin{figure}[p]
    \centering
    \includegraphics[width=\linewidth,height=0.76\textheight,keepaspectratio]{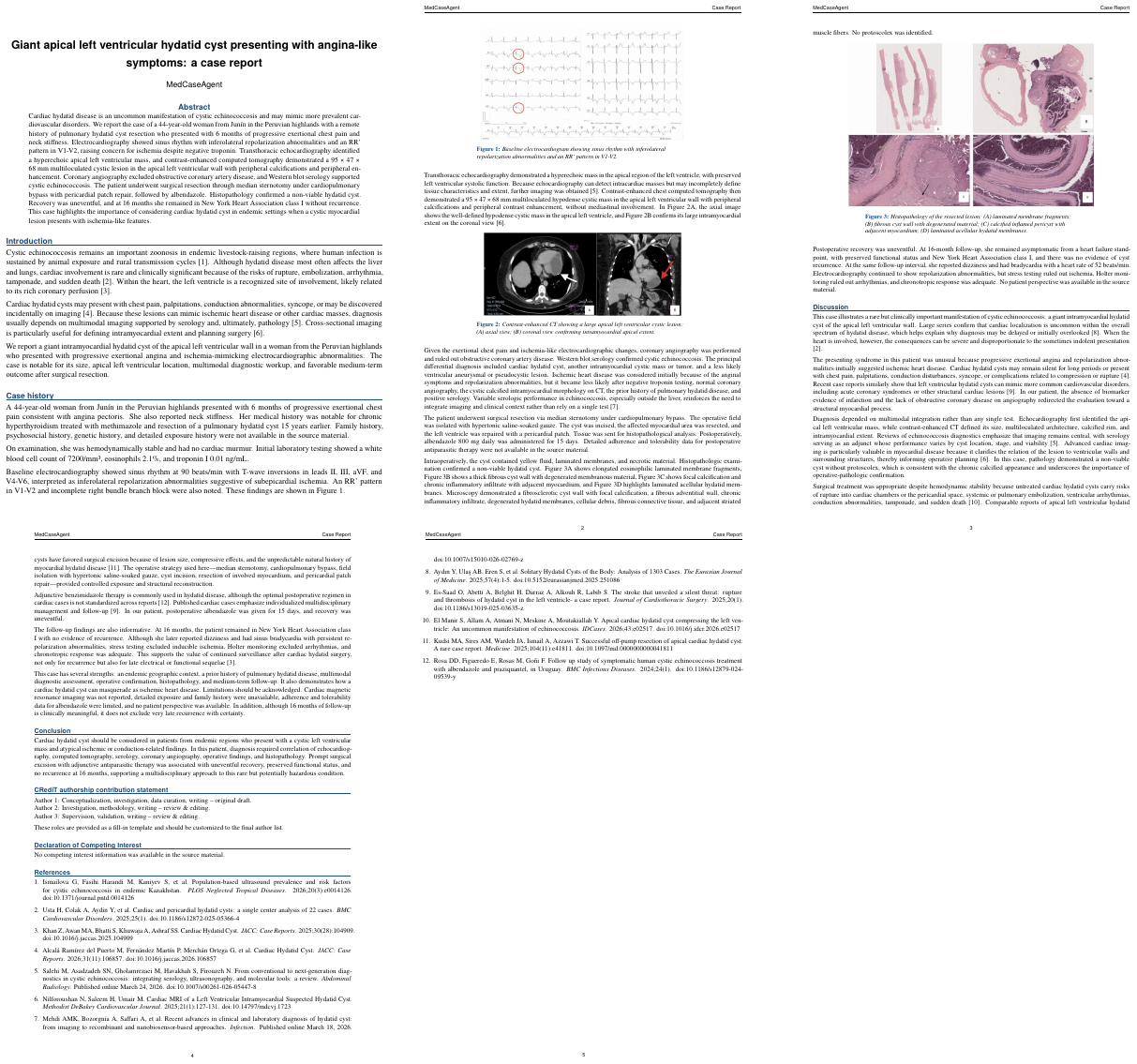}
    \caption{Example 3 of a case report generated by MedCaseAgent. The full-page sheet shows the generated manuscript structure, image placement, citation formatting, and clinical timeline organization.}
    \label{fig:appendix_generated_report_sheet_3}
\end{figure}

Figures~\ref{fig:appendix_generated_report_sheet_1}--\ref{fig:appendix_generated_report_sheet_3} show three complete example case reports generated by MedCaseAgent. Each generated output is included as a full-page sheet to show the manuscript structure, medical-image placement, citation formatting, and clinical timeline organization. These examples are qualitative outputs from the planner--writer--editor pipeline and complement the quantitative results in the main paper.

\FloatBarrier
\clearpage

\section{Broader Impacts}
\label{subsec:broader_impacts}
\setcounter{figure}{0}
\setcounter{table}{0}

MedCaseAgent is intended to support clinicians in writing case reports. Case-report writing is a time-consuming part of clinical knowledge sharing: clinicians must organize patient history, diagnostic evidence, images, treatment course, outcomes, and related literature before a report can be submitted. Unlike open-ended research-paper generation, this writing task is strongly evidence-based and grounded in an observed patient case, making it well suited to a clinician-controlled tool that structures evidence rather than inventing new claims.

The main positive impact is therefore to reduce the documentation and literature-review burden involved in manuscript preparation. MedCaseAgent helps organize fragmented notes, medical images, prior-case evidence, and citation candidates into a structured draft that remains available for expert review. This support is especially useful for rare, atypical, or unprecedented presentations, where clinicians must compare a complex patient course against sparse prior literature.

MedCaseAgent may also provide useful cues before writing begins. During the planning stage, retrieval over analogous cases, literature, images, and disease-level evidence can help surface diagnostic possibilities or comparative context that clinicians may want to consider. These planning outputs are not meant to replace expert diagnosis, but they can serve as organized signals for specialists who are already evaluating a difficult case. More broadly, the system may enable faster dissemination of rare-disease observations, more consistent case-report structure, and broader participation from clinical teams with limited time for manuscript drafting.

Responsible use should keep clinicians at the center of the workflow. Patient facts, image interpretations, and citation support should be verified before submission or dissemination, and any deployment on real clinical records should follow local governance, consent, de-identification, and access-control requirements.

\FloatBarrier
\clearpage

\section{Release and LLM Usage Disclosure}
\label{subsec:release_llm_disclosure}
\setcounter{figure}{0}
\setcounter{table}{0}

We will release the MedCase-150K and MedCase-Bench datasets, together with the MedCaseAgent source code, data-construction scripts, evaluation prompts, metadata, and data splits, under CC-BY-NC 4.0.

This disclosure concerns LLMs as the modeling technology studied in this work, including the agentic planner, writer, editor, and LLM-judge evaluation components described in the main paper. Separate from the method, LLM-based tools were used only for manuscript writing and editing assistance. They were not used to fabricate patient records, experimental measurements, baselines, or evaluation outcomes; final reported numbers must be produced by the evaluation pipeline and be traceable to released scripts, prompts, and logs.

\clearpage

\end{document}